\documentclass{article}
\let\standardaddcontentsline\addcontentsline
\usepackage[T1]{fontenc}
\usepackage{colm2024_conference}
\let\addcontentsline\standardaddcontentsline

\usepackage{booktabs}
\usepackage{graphicx}
\usepackage{enumitem}
\usepackage{needspace}
\usepackage{wrapfig}
\usepackage{algorithm}
\usepackage{algpseudocode}
\usepackage{wrapfig}
\usepackage{float}
\usepackage{placeins}
\usepackage{microtype}
\usepackage{amsmath}
\usepackage{amssymb}
\usepackage{mathtools}
\usepackage{colortbl}
\usepackage[utf8]{inputenc}
\usepackage{caption}
\usepackage{subcaption}
\usepackage{xcolor}
\usepackage{setspace}
\usepackage{url}
\usepackage{multirow}
\usepackage{colortbl}
\usepackage{tabularx}
\usepackage{blindtext}
\makeatletter
\renewcommand{\blindtext}[1][\value{blindtext}]{%
  \begingroup
  \color{red}%
  \blind@checklanguage
  \setcounter{blind@randommax}{#1}%
  \setcounter{blind@pangrammax}{#1}%
  \blind@countxx=1 %
  \loop
    \blindtext@text\
  \ifnum\blind@countxx<#1\advance\blind@countxx by 1 %
  \repeat
  \endgroup
}
\makeatother
\usepackage{pgfplots}
\pgfplotsset{compat=1.18}
\usepackage{tikz}
\usepackage[edges]{forest}
\usepackage{ragged2e}
\usetikzlibrary{er, positioning, bayesnet, arrows.meta, calc}
\usepackage{makecell}
\usepackage{siunitx}
\usepackage{nicefrac}
\usepackage{tocloft}
\usepackage{etoc}
\usepackage{listings}
\usepackage[raster, skins]{tcolorbox}
\usepackage{xltabular}
\usepackage{adjustbox}
\usepackage{xurl}
\usepackage{xspace}
\usepackage{longtable}
\usepackage{pifont}
\usepackage{fontawesome5}

\hypersetup{
  linkcolor=citationblue,
  citecolor=citationblue,
  urlcolor=citationblue
}

\usepackage{amsmath,amsfonts,bm}

\def\eqref#1{equation~\ref{#1}}

\def\1{\bm{1}}

\DeclareMathAlphabet{\mathsfit}{\encodingdefault}{\sfdefault}{m}{sl}
\SetMathAlphabet{\mathsfit}{bold}{\encodingdefault}{\sfdefault}{bx}{n}

\definecolor{lightgray}{rgb}{0.9,0.9,0.9}
\definecolor{catgray}{HTML}{F0F0F0}
\definecolor{codewindowbg}{HTML}{FFFFFF}
\definecolor{codewindowbar}{HTML}{F3F3F3}
\definecolor{codewindowframe}{HTML}{B8BDC3}
\definecolor{codewindowtext}{HTML}{2D3136}
\definecolor{codewindowmuted}{HTML}{555B63}
\definecolor{codewindowteal}{HTML}{247B73}
\definecolor{codewindowblue}{HTML}{2F6DB5}
\definecolor{jsonbraceouter}{HTML}{FF6B6B}
\definecolor{jsonbracearray}{HTML}{FF9F0A}
\definecolor{jsonbraceobject}{HTML}{9B6DFF}
\definecolor{jsonbraceinner}{HTML}{22B8CF}
\definecolor{tracewave}{HTML}{C65A11}
\definecolor{codewindowred}{HTML}{FF5F57}
\definecolor{codewindowyellow}{HTML}{FEBC2E}
\definecolor{codewindowgreen}{HTML}{28C840}

\lstdefinestyle{appendixcode}{
  basicstyle=\footnotesize\ttfamily,
  backgroundcolor=\color{catgray},
  frame=single,
  rulecolor=\color{black!35},
  framerule=0.4pt,
  framesep=5pt,
  xleftmargin=0.5em,
  xrightmargin=0.5em,
  framexleftmargin=4pt,
  framexrightmargin=4pt,
  framextopmargin=4pt,
  framexbottommargin=4pt,
  breaklines=true,
  breakatwhitespace=false,
  columns=fullflexible,
  keepspaces=true,
  showstringspaces=false,
  tabsize=2,
  captionpos=t,
  aboveskip=10pt,
  belowskip=10pt
}

\lstdefinestyle{skillcode}{
  basicstyle=\footnotesize\ttfamily\color{codewindowtext},
  morecomment=[l]{\#},
  commentstyle=\color{codewindowteal}\bfseries,
  morestring=[b]",
  stringstyle=\color{codewindowblue},
  alsoletter={._},
  morekeywords={generate.image,generate_image,node.prompt,node.params},
  keywordstyle=\color{codewindowblue},
  breaklines=true,
  breakatwhitespace=false,
  columns=fullflexible,
  keepspaces=true,
  showstringspaces=false,
  tabsize=2,
  aboveskip=0pt,
  belowskip=0pt
}

\lstdefinestyle{jsoncode}{
  style=skillcode,
  basicstyle=\footnotesize\ttfamily\color{codewindowmuted},
  stringstyle=\color{codewindowblue},
  escapeinside={(*@}{@*)}
}
\newcommand{\jsonkeytoken}[1]{\textcolor{codewindowteal}{\ttfamily #1}}
\newcommand{\jsonoutertoken}[1]{\textcolor{jsonbraceouter}{\ttfamily #1}}
\newcommand{\jsonarraytoken}[1]{\textcolor{jsonbracearray}{\ttfamily #1}}
\newcommand{\jsonobjecttoken}[1]{\textcolor{jsonbraceobject}{\ttfamily #1}}
\newcommand{\jsoninnertoken}[1]{\textcolor{jsonbraceinner}{\ttfamily #1}}

\newcommand{\jsonturnline}[1]{%
  \makebox[\linewidth][l]{%
    \textcolor{skillpurple}{\ttfamily\bfseries
      \hbox to 1.8em{\leaders\hbox{\char45}\hfill}\hspace{0.45em}%
      #1\hspace{0.45em}%
      \leaders\hbox{\char45}\hfill\kern0pt}}%
}

\definecolor{skillblue}{HTML}{2F6DB5}
\definecolor{skillgreen}{HTML}{2E8B57}
\definecolor{skillpurple}{HTML}{6A4C93}

\definecolor{levelatomic}{HTML}{599BD4}
\definecolor{levelexpert}{HTML}{0ED1D2}
\definecolor{levelscenario}{HTML}{70AE47}

\definecolor{archentry}{HTML}{C65A11}
\definecolor{archmcp}{HTML}{B58801}
\definecolor{archprovider}{HTML}{70AE47}
\definecolor{archregistry}{HTML}{5757AF}

\definecolor{coverageNavy}{RGB}{20,68,106}
\definecolor{coverageReferenceRed}{RGB}{205,44,36}
\definecolor{coverageReferenceYellow}{RGB}{255,228,123}
\definecolor{coverageReferenceGreen}{RGB}{34,139,34}
\definecolor{coverageReferenceBlue}{RGB}{194,232,247}
\colorlet{coverageOrangeFill}{coverageReferenceRed!20}
\definecolor{coverageOrangeFrame}{HTML}{C65A11}
\colorlet{coverageYellowFill}{coverageReferenceYellow!32}
\definecolor{coverageYellowFrame}{HTML}{B58801}
\colorlet{coverageGreenFill}{coverageReferenceGreen!20}
\definecolor{coverageGreenFrame}{HTML}{70AE47}
\colorlet{coveragePurpleFill}{coverageReferenceBlue!57}
\definecolor{coveragePurpleFrame}{HTML}{5757AF}

\definecolor{quotebg}{HTML}{EEF2FD}
\definecolor{quoteborder}{HTML}{2563EB}

\definecolor{takeawaybg}{HTML}{F0F0FF}
\definecolor{takeawayframe}{HTML}{000000}

\tcbuselibrary{breakable,listings}
\newcommand{\windowdot}[1]{%
  \tikz[baseline=-0.62ex]{%
    \filldraw[fill=#1,draw=#1!78!black,line width=0.22pt]
      (0,0) circle (2.65pt);}%
}
\newcommand{\windowcontrols}{%
  \windowdot{codewindowred}\hspace{4.5pt}%
  \windowdot{codewindowyellow}\hspace{4.5pt}%
  \windowdot{codewindowgreen}%
}
\newtcblisting{skillfigurebox}[1]{%
  enhanced jigsaw,
  listing only,
  listing engine=listings,
  listing options={style=skillcode},
  colback=codewindowbg,
  colframe=codewindowframe,
  colbacktitle=codewindowbar,
  coltitle=codewindowmuted,
  boxrule=0.9pt,
  arc=5pt,
  outer arc=5pt,
  left=10pt, right=10pt, top=6.5pt, bottom=6.5pt,
  boxsep=0pt,
  before skip=8.5pt, after skip=8.5pt,
  fonttitle=\footnotesize\sffamily,
  lefttitle=10pt,
  righttitle=10pt,
  toptitle=5.1pt,
  bottomtitle=5.1pt,
  titlerule=0.7pt,
  title={%
    \windowcontrols
    \hfill\textcolor{codewindowtext}{\texttt{\bfseries #1}}\hfill
    \phantom{\windowcontrols}%
  }
}
\newtcblisting{jsonlistingbox}[1]{%
  enhanced jigsaw,
  listing only,
  listing engine=listings,
  listing options={style=jsoncode},
  colback=codewindowbg,
  colframe=codewindowframe,
  colbacktitle=codewindowbar,
  coltitle=codewindowmuted,
  boxrule=0.9pt,
  arc=5pt,
  outer arc=5pt,
  left=10pt, right=10pt, top=6.5pt, bottom=6.5pt,
  boxsep=0pt,
  before skip=8.5pt, after skip=8.5pt,
  fonttitle=\footnotesize\sffamily,
  lefttitle=10pt,
  righttitle=10pt,
  toptitle=5.1pt,
  bottomtitle=5.1pt,
  titlerule=0.7pt,
  title={%
    \windowcontrols
    \hfill\textcolor{codewindowtext}{\texttt{\bfseries #1}}\hfill
    \phantom{\windowcontrols}%
  }
}
\newtcolorbox{takeaway}[3][Takeaways]{%
  enhanced, breakable,
  colback=#2,
  colframe=#3,
  coltitle=white,
  fonttitle=\bfseries,
  title={\faLightbulb[regular]~#1},
  attach boxed title to top left={xshift=8pt,yshift*=-\tcboxedtitleheight/2},
  boxed title style={%
    colback=#3,
    colframe=#3,
    boxrule=0pt,
    arc=2pt,
    left=6pt, right=6pt, top=2pt, bottom=2pt,
  },
  boxrule=0.9pt,
  arc=4.25pt,
  left=11pt, right=11pt, top=8pt, bottom=7pt,
  boxsep=0pt,
  before skip=13pt, after skip=10pt,
}
\newtcolorbox{epigraph}[1]{%
  enhanced,
  colback=quotebg,
  colframe=quoteborder,
  boxrule=0.9pt,
  arc=3pt,
  left=8pt, right=8pt, top=6pt, bottom=5pt,
  boxsep=1pt,
  before skip=4pt, after skip=10pt,
  fontupper=\small\bfseries\itshape,
  after upper={\par\vspace{3pt}\raggedleft\small\bfseries\itshape #1\par},
}

\newlist{tkitems}{itemize}{1}
\setlist[tkitems]{
  label=\textbullet,
  leftmargin=18pt, labelsep=6pt,
  topsep=0pt, itemsep=3pt, parsep=0pt, partopsep=0pt,
}

\newtcolorbox{skillbox}[2]{%
  enhanced, breakable,
  colback=#1!4!white,
  colframe=#1,
  coltitle=white,
  fonttitle=\bfseries,
  title={#2},
  boxrule=0.7pt,
  arc=2pt,
  left=6pt, right=6pt, top=5pt, bottom=5pt,
  boxsep=2pt,
  before skip=9pt, after skip=9pt,
}

\renewcommand{\arraystretch}{1.25}

\makeatletter
\DeclareRobustCommand\onedot{\futurelet\@let@token\@onedot}
\def\@onedot{\ifx\@let@token.\else.\null\fi\xspace}

\makeatother

\urldef{\ghlinknew}\url{https://github.com/any2any-mllm/Omni-IO-Skill}

\title{
Omni-IO Skills: Harnessing Your Agent Omni-Native
}

\author{
Yanlin Li $^{1,\dagger}$
\,
Mingyang Hao $^{1}$
\,
Shengqiong Wu $^{2}$
\,
Hao Fei $^{2,\ddagger}$
\,
Mong-Li Lee $^{1}$
\,
Wynne Hsu $^{1}$
\\
{\normalfont\small
 $^{1}$ National University of Singapore
 \quad
 $^{2}$ University of Oxford}
}

\newcommand{\cmark}{{\color{green!70!black}\ding{51}}}
\newcommand{\xmark}{{\color{red!80!black}\ding{55}}}

\makeatletter
\def\@maketitle{%
  \vbox{\hsize\textwidth
    {\centering {\Large\bf \@title\par}}
    \vspace{-0.1cm}
    \colmfullwidthauthors
    \vskip 0.18in minus 0.05in
  }%
  \begin{center}
    \vspace{-1cm}
    {\ttfamily\bf Code Repo: \ghlinknew}
  \end{center}
  \thispagestyle{firstpage}
}
\makeatother

\begin{document}
\maketitle
\begingroup
\renewcommand{\thefootnote}{\fnsymbol{footnote}}
\footnotetext[2]{Email: yanlin.li@u.nus.edu.}
\footnotetext[3]{Correspondence, project lead: Hao Fei.
Email: haofei7419@gmail.com.}
\endgroup

\vspace{0.4cm}
\begin{abstract}

General-purpose agents can plan, reason, and act over long horizons, yet their production capabilities remain fragmented across text, images, audio, video, documents, 3D assets, and code.
Extending a foundation model to additional modalities ties capability growth to costly model updates, while assembling specialist models and tools leaves unresolved how procedures, dependencies, intermediate assets, and cross-turn revisions should be coordinated.
We present \textbf{Omni-IO Skills}, a plug-and-play Agent Harness that makes existing agents omni-native through hierarchical Skills, a standardized multimodal execution interface, dependency-aware orchestration, and a persistent Asset Registry.
Multi-asset workflows are represented as Declare Execution Graphs, which schedule independent operations concurrently and register successful outputs for downstream and cross-turn reuse across replaceable execution backends.
Its 27 Skills cover 38 representative tasks spanning seven artifact modalities and four capability families: understanding, generation, reasoning, and retrieval.
On UniM-90, the harness raises the input-support rates of GPT-5.6 Sol and Claude Sonnet 5 from 40.00\% and 38.89\% to 100\%, while increasing relative Semantic--Quality Coupled Score from 26.99 to 74.94 and from 27.82 to 77.78, respectively; Strict Structure Score reaches 100.00 and 99.78.
These results establish harness-level capability composition as a practical route to broad, evolvable Omni systems without changing the host agent's reasoning core.

\end{abstract}

\begin{figure}[!h]
  \centering
  \includegraphics[width=\linewidth]{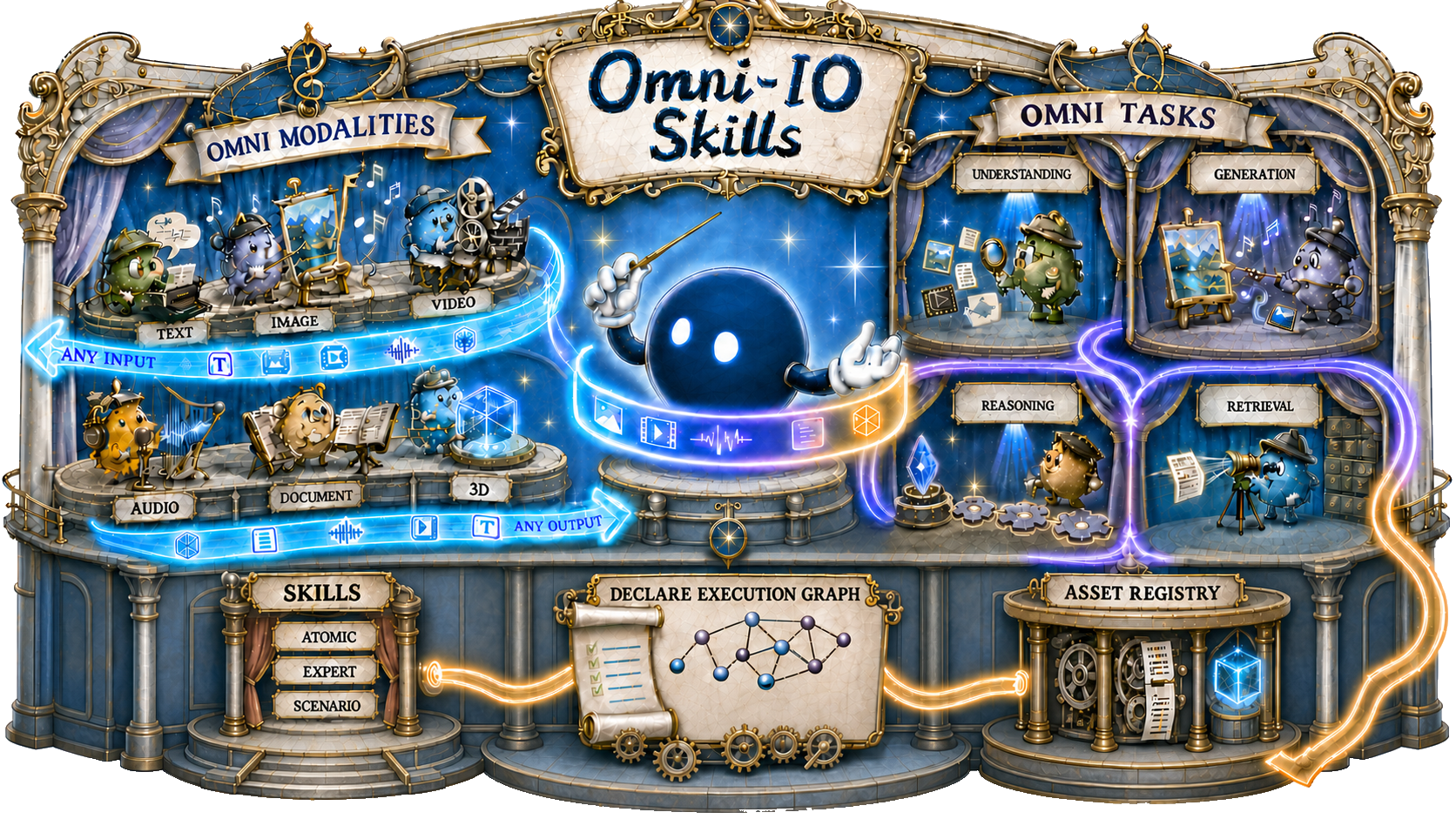}
  \caption{Overview of Omni-IO Skills.
  Hierarchical Skills orchestrate multimodal understanding and generation, while generated assets are registered for downstream and cross-turn reuse.}
  \label{fig:teaser}
\end{figure}
\etocdepthtag.toc{mainmatter}
\section{Introduction}

Real-world tasks rarely remain within a single modality.
Producing an online course, for example, may begin with lecture recordings and reference documents, continue through content analysis and visual design, and end with slides, illustrations, narration, and an explainer video.
The resulting artifacts are not independent outputs: they share facts, style, timing, and production constraints.
An effective Omni system must therefore receive and produce interleaved combinations of text, images, audio, video, documents, 3D assets, and code while preserving semantic and asset continuity across the entire workflow~\cite{li2026unim}.
Recent Omni foundation models have expanded native understanding and generation through unified autoregressive modeling~\cite{unifiedio2,anygpt,chameleon} and hybrid discrete--continuous designs~\cite{nextgpt,transfusion}.
This progress also exposes a persistent scaling pressure.
As more modalities enter a shared model, their representations, objectives, and fidelity requirements must be reconciled, often with new data, codecs, decoders, and alignment stages~\cite{janus}.
At the same time, specialized models and media engines continue to improve outside the shared backbone.
The resulting ecosystem motivates a complementary route to Omni capability: a system layer that can harness heterogeneous and continuously evolving capabilities into coherent, application-level workflows.

General-purpose agents such as Codex and Claude Code already provide strong instruction following, long-horizon planning, code execution, workspace operation, and iterative refinement~\cite{openai2026codexapp,anthropic2025claudecode}.
They supply much of the reasoning core needed to pursue complex goals, yet their end-to-end production surface remains centered on software and knowledge work.
Tasks involving audio, video, 3D, professional documents, and coordinated media packages still depend on external capabilities and explicit orchestration.
Model-coordination systems show that an agent can delegate to modality specialists without retraining its underlying model~\cite{agentomni}, while recent Agent Skills package reusable procedural knowledge for inference-time use~\cite{skillsbench,cuaskill,mmskills}.
These two ingredients do not by themselves provide an Omni workflow.
A production task must also select the right capabilities, express control and data dependencies, move intermediate artifacts across tools, isolate local failures, and recover prior outputs for subsequent turns.
The missing component is an \emph{Omni agent harness}: a layer between agent reasoning and heterogeneous execution backends that turns scattered models, tools, and procedures into selectable, composable, executable, and traceable capabilities.
This leads to our central question: \textit{can a plug-and-play harness make an existing general-purpose agent omni-native while preserving its reasoning and planning core?}

We introduce \textbf{Omni-IO Skills}, a plug-and-play Omni-modal agent harness whose capabilities are carried by loadable Skills.
As illustrated in Figure~\ref{fig:teaser}, the harness leaves the host agent unchanged and provides the procedural knowledge, execution interface, and persistent asset substrate required to turn a user request into a complete multimodal workflow.
Its Atomic, Expert, and Scenario Skills capture reusable capability primitives, production procedures for concrete deliverables, and application-level tasks with multiple related outputs.
The current implementation comprises 19 Atomic, 2 Expert, and 6 Scenario Skills.
Together, these 27 Skills cover 38 representative tasks across understanding, generation, reasoning, and retrieval, spanning seven artifact modalities and application domains from education and research to marketing, creative production, and software engineering; Section~\ref{sec:capabilities} presents this user-facing scope.
Underneath the Skills, a layered architecture separates task knowledge, tool access, provider binding, and artifact state.
Dependency-aware execution and a persistent Asset Registry then coordinate task composition, artifact transfer, and later revision without binding a workflow to a particular service or file path; Sections~\ref{sec:system_design} and~\ref{sec:orchestration} detail these mechanisms.
Coupled with this runtime and asset layer, the Skills form an Agent Harness that can evolve with its models, tools, and application requirements.

We evaluate the harness on UniM-90, a controlled 90-instance subset of UniM~\cite{li2026unim}, using GPT-5.6 Sol and Claude Sonnet 5 as two distinct host agents~\cite{openai2026gpt56,anthropic2026sonnet5}.
Omni-IO Skills raises their input-support rates from 40.00\% and 38.89\% to 100\%.
Across the full test set, relative Semantic--Quality Coupled Score (SQCS) increases from 26.99 to 74.94 for GPT-5.6 Sol and from 27.82 to 77.78 for Claude Sonnet 5, gains of 47.95 and 49.96 percentage points.
Strict Structure Score reaches 100.00 and 99.78, respectively, while both hosts attain a Lenient Structure Score of 100.00.
The absolute SQCS values after loading the harness are 74.94 and 77.78 over all instances, exceeding the Base Agents' 67.49 and 71.53 measured only on their narrower supported subsets.
These results show that a shared harness can close substantial modality and workflow gaps across different hosts while delivering strong response quality on all 90 instances.

Our contributions are fourfold:
\begin{itemize}[leftmargin=1.4em,itemsep=2pt,topsep=3pt,parsep=0pt]
  \item We formulate an Omni-modal Agent Harness that extends a general-purpose agent into an omni-native system without retraining the host model.

  \item We realize this formulation as Omni-IO Skills, integrating multi-granularity procedural knowledge, heterogeneous execution backends, dependency-aware orchestration, and persistent artifact state within one extensible architecture.

  \item We build an application-facing system that covers seven artifact modalities, 27 Skills, and 38 representative tasks across understanding, generation, reasoning, and retrieval.

  \item We demonstrate consistent gains in modality coverage, semantic quality, interleaved coherence, and structural completeness across two different host agents.
\end{itemize}

\section{Related Work}

\subsection{Omni Foundation Models}

Early research on multimodal understanding and generation follows largely separate paths.
Understanding models map images, audio, and video into semantic representations for language-based inference~\cite{clip,liu2026interference,videollama,luo2024panosent,lin2026towards}, whereas generative models specialize in recovering visual, acoustic, or temporal detail~\cite{dalle2,audiolm,li2025vegas,videodiffusion,liu2025gaming,luo2026unveiling}.
Omni foundation models have since brought perception, reasoning, and generation into a shared context.
Their architectures broadly follow two routes.
Unified autoregressive models discretize heterogeneous modalities and apply next-token prediction over a common sequence, providing a uniform interface for mixed and interleaved content~\cite{unifiedio2,anygpt,chameleon}.
Hybrid discrete-continuous models retain autoregressive semantic reasoning while using continuous representations, diffusion or flow objectives, or modality-specific decoders to preserve media fidelity~\cite{nextgpt,transfusion,janusflow,qwen25omni}.
Autoregressive unification simplifies the architecture and naturally supports interleaving, at the cost of modality tokenization, sequence length, and perceptual-detail pressure.
Hybrid designs retain greater specialization and must coordinate multiple encoders, adapters, objectives, and decoders.
These routes form a design spectrum, with recent systems also introducing streaming, full-duplex interaction, and modality-specific experts to reduce latency and cross-modal interference~\cite{moshi,qwen3omni,mingomni}.
Such advances expand Omni capability within the model, whose supported modalities and outputs remain coupled to its training data, architecture, and update cycle.
Omni-IO Skills is complementary: it treats foundation models, specialist models, and media engines as replaceable execution backends, and places unification at the level of task execution and artifact flow so that applications can evolve independently of a particular model stack.

\subsection{Agent Harnesses and Skills}

A foundation model supplies a reasoning and decision policy; an Agent Harness provides the operational substrate for sustained execution, including the action loop, tool access, context and state management, execution control, verification, and recovery~\cite{agentharnesssurvey,luo2026dr,wu2026safeguard}.
ReAct establishes an interleaved reasoning-action-observation loop through which a model can revise its plan from environmental feedback~\cite{react}, while the Model Context Protocol standardizes how applications expose external tools and resources to models~\cite{mcp}.
These mechanisms define an agent's action surface.
Reliable long-horizon execution additionally depends on how the system carries state, represents dependencies, verifies results, and contains failures.
Multimodal agents use similar control loops to select and coordinate modality specialists: MM-ReAct connects a language model to vision experts~\cite{mmreact}, and recent Omni agents extend coordination to images, audio, and video through master-agent delegation or active perception~\cite{agentomni,omnigaia}.
These systems are commonly evaluated on evidence acquisition, question answering, cross-modal reasoning, and response integration.
Production workflows that create multiple dependent artifacts further require intermediate-asset transfer, provenance, and cross-turn revision.
Omni-IO Skills extends the harness along these dimensions, connecting a general-purpose agent to heterogeneous multimodal backends through dependency-aware execution and persistent artifact state.

Agent Skills add a procedural knowledge layer to the harness.
A Skill records when a task applies, which inputs it requires, how tools should be invoked or composed, and what outputs should be produced.
It can therefore preserve tested workflows, domain conventions, executable code, and composition patterns beyond the atomic operations exposed by tools.
SkillsBench measures the effect of curated Skills across diverse expert tasks~\cite{skillsbench}; CUA-Skill represents computer-use procedures with parameterized execution and composition graphs~\cite{cuaskill}; and MMSkills couples textual procedures with state cards and visual keyframes for visual decision making~\cite{mmskills}.
Skills now span general expert tasks, computer use, and visual agents, while Omni-agent research separately establishes the value of coordinating modality experts.
Their intersection remains underdeveloped for workflows that combine any-to-any understanding and generation with multi-asset execution and persistent artifact state.
Omni-IO Skills connects these lines: hierarchical Skills organize multimodal procedures, and the surrounding Harness turns them into composable, traceable workflows with persistent cross-turn asset reuse.

\section{System Capabilities and Task Coverage}
\label{sec:capabilities}

Omni-IO Skills provides an application-facing task interface for requests whose source material, intermediate assets, and deliverables span multiple media types.
A request may begin with a report, a recorded interview, product images, or an existing 3D model, then produce an analysis, a new media artifact, or a coordinated package of outputs.
Figure~\ref{fig:task_coverage} presents this user-visible capability surface across four operation families and the domains in which they are commonly applied.

\paragraph{Modality Coverage.}
The implemented Skills accept, produce, and connect seven artifact types:
\begin{center}
  \begingroup
  \small
  \setlength{\tabcolsep}{4pt}
  \begin{tabular}{@{}ccccccc@{}}
    \mbox{\raisebox{-0.18\height}{\includegraphics[height=1.05em]{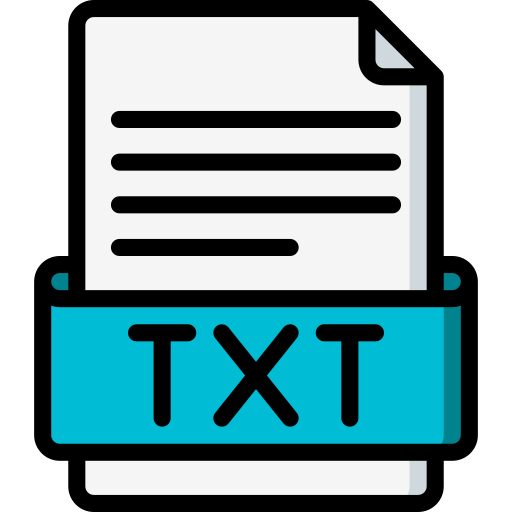}}\hspace{2pt}\textbf{Text}} &
    \mbox{\raisebox{-0.18\height}{\includegraphics[height=1.05em]{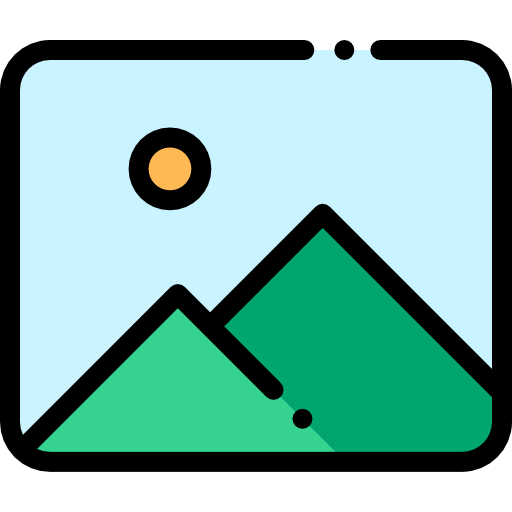}}\hspace{2pt}\textbf{Image}} &
    \mbox{\raisebox{-0.18\height}{\includegraphics[height=1.05em]{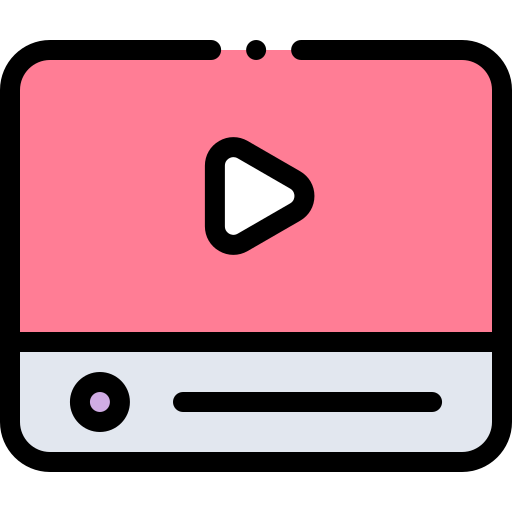}}\hspace{2pt}\textbf{Video}} &
    \mbox{\raisebox{-0.18\height}{\includegraphics[height=1.05em]{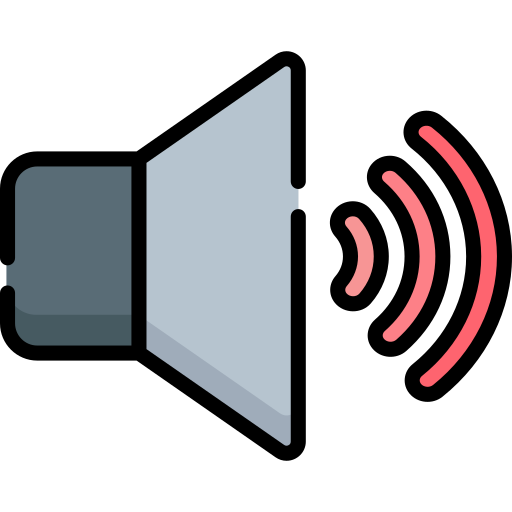}}\hspace{2pt}\textbf{Audio}} &
    \mbox{\raisebox{-0.18\height}{\includegraphics[height=1.05em]{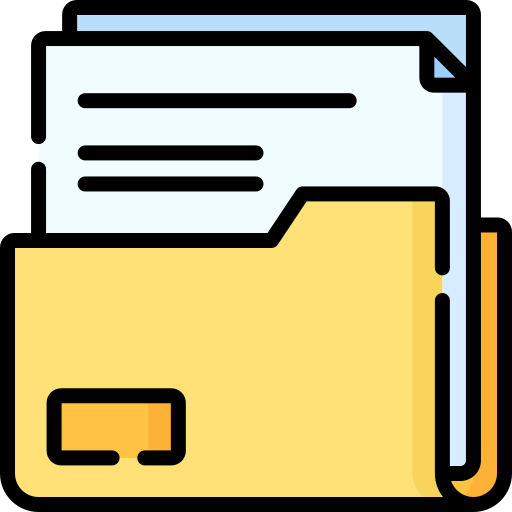}}\hspace{2pt}\textbf{Document}} &
    \mbox{\raisebox{-0.18\height}{\includegraphics[height=1.05em]{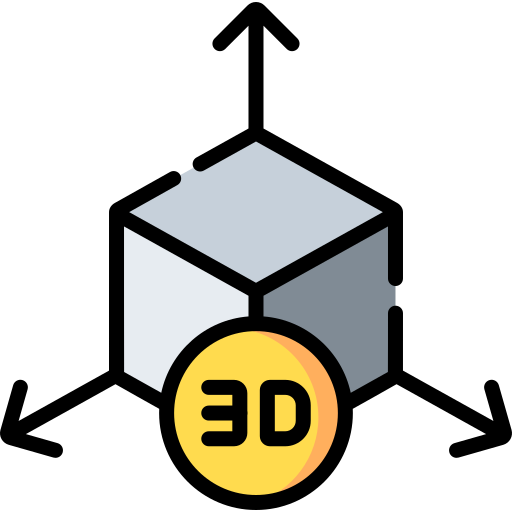}}\hspace{2pt}\textbf{3D}} &
    \mbox{\raisebox{-0.18\height}{\includegraphics[height=1.05em]{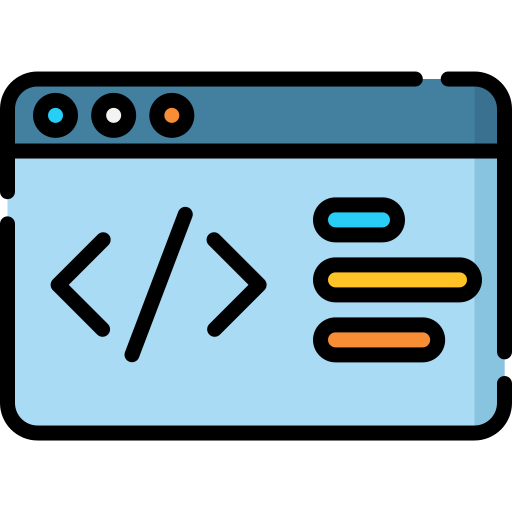}}\hspace{2pt}\textbf{Code}}
  \end{tabular}
  \endgroup
\end{center}
The same visual vocabulary is used in Figure~\ref{fig:task_coverage}.
An icon on the left of an arrow denotes an input, while an icon on the right denotes an output.
Several icons on either side indicate a task that consumes or produces multiple artifact types.

\tikzset{
  coverage box/.style={
    rectangle,
    draw=coverageNavy,
    rounded corners=2pt,
    text=black,
    line width=0.9pt,
    inner xsep=5pt,
    inner ysep=4pt,
    align=center,
  },
  coverage category/.style={
    coverage box,
    fill=white,
    font=\normalsize\bfseries,
  },
  coverage function/.style={
    coverage box,
    fill=white,
    draw=coverageNavy,
    font=\normalsize\bfseries\itshape,
    inner xsep=5pt,
    inner ysep=4pt,
  },
  coverage panel/.style={
    coverage box,
    align=left,
    font=\normalsize,
    inner xsep=7pt,
    inner ysep=6pt,
  },
  coverage understanding/.style={coverage panel, fill=coverageOrangeFill, draw=coverageNavy},
  coverage generation/.style={coverage panel, fill=coverageYellowFill, draw=coverageNavy},
  coverage reasoning/.style={coverage panel, fill=coverageGreenFill, draw=coverageNavy},
  coverage retrieval/.style={coverage panel, fill=coveragePurpleFill, draw=coverageNavy},
}

\newcommand{\coveragefacet}[3]{%
  \begin{tabular}[t]{@{}!{\color{#1}\vrule width 1.2pt\hspace{4pt}}p{16.3em}@{}}
    \textcolor{#1}{\textit{\textbf{#2}}}\par\vspace{1pt}%
    {\color{black}\small\RaggedRight #3\par}
  \end{tabular}%
}

\newcommand{\coveragetasks}[1]{%
  \parbox[t]{28em}{\small\RaggedRight \textit{e.g.,} #1, \textit{etc.}\par}%
}

\newcommand{\coveragefunctionlabel}[1]{%
  \parbox[c]{12.5em}{\centering #1}%
}

\newcommand{\coverageindustrytasks}[1]{%
  \parbox[t]{27.75em}{\small\RaggedRight \textit{e.g.,} #1, \textit{etc.}\par}%
}

\newcommand{\coverageModText}{%
  \raisebox{-0.18\height}{\includegraphics[height=0.9em]{logo/text.png}}%
}
\newcommand{\coverageModImage}{%
  \raisebox{-0.18\height}{\includegraphics[height=0.9em]{logo/image.png}}%
}
\newcommand{\coverageModAudio}{%
  \raisebox{-0.18\height}{\includegraphics[height=0.9em]{logo/audio.png}}%
}
\newcommand{\coverageModVideo}{%
  \raisebox{-0.18\height}{\includegraphics[height=0.9em]{logo/video.png}}%
}
\newcommand{\coverageModDocument}{%
  \raisebox{-0.18\height}{\includegraphics[height=0.9em]{logo/document.png}}%
}
\newcommand{\coverageModCode}{%
  \raisebox{-0.18\height}{\includegraphics[height=0.9em]{logo/code.png}}%
}
\newcommand{\coverageModThreeD}{%
  \raisebox{-0.18\height}{\includegraphics[height=0.9em]{logo/3d.png}}%
}

\newcommand{\coverageModSep}{\hspace{1.5pt}}
\newcommand{\coverageModFlow}[2]{%
  \nobreakspace\mbox{(#1\hspace{2pt}$\rightarrow$\hspace{2pt}#2)}%
}

\newcommand{\coverageindustrylabel}[1]{%
  \parbox[c]{11.5em}{\centering #1}%
}

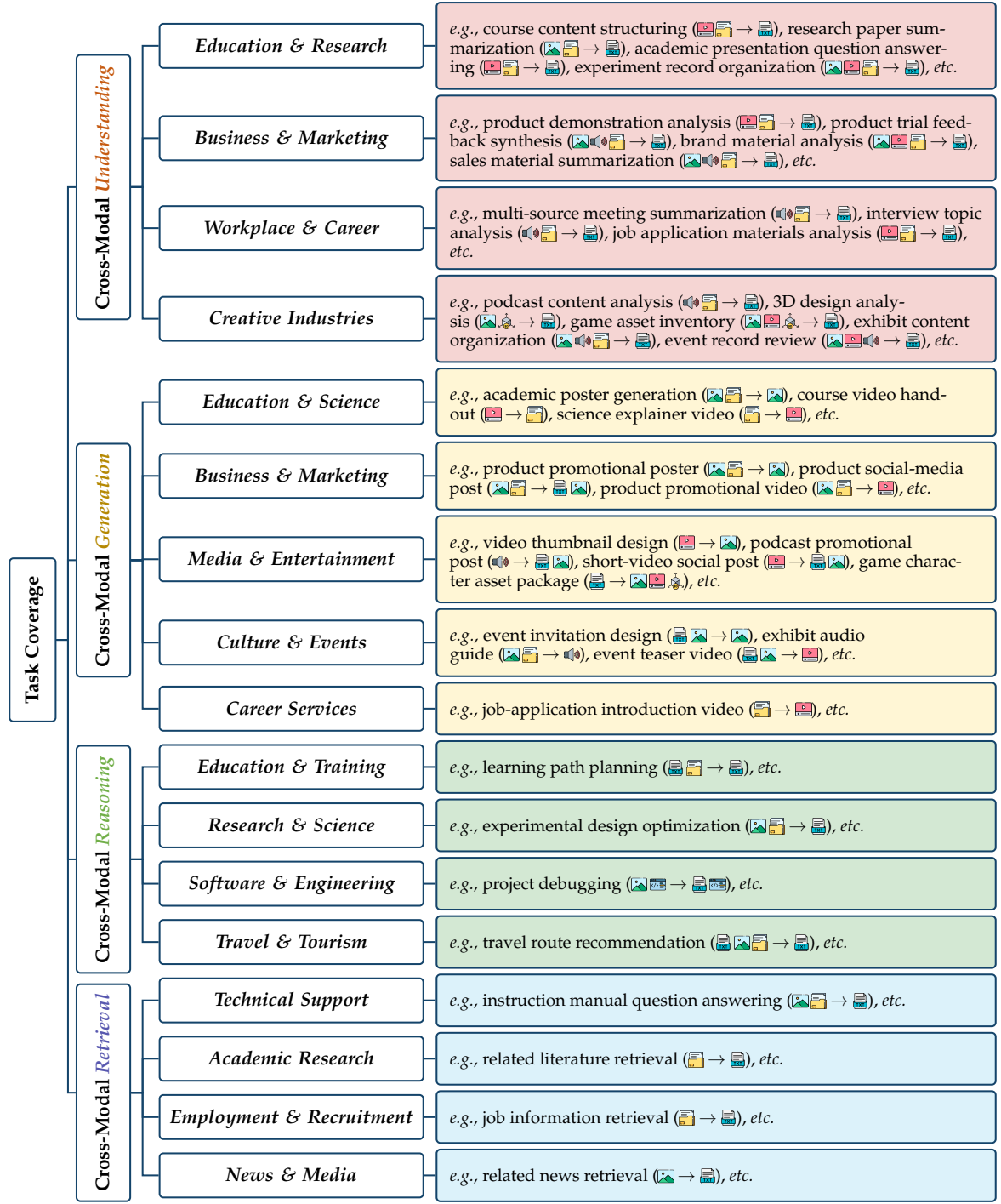
\begin{figure}[!t]
  \centering
  \resizebox{0.96\textwidth}{!}{%
    \begin{forest}
      forked edges,
      for tree={
        grow=east,
        reversed=true,
        anchor=base west,
        parent anchor=east,
        child anchor=west,
        base=left,
        rectangle,
        draw=coverageNavy,
        rounded corners=2pt,
        align=center,
        font=\normalsize,
        edge+={coverageNavy, line width=0.9pt},
        s sep=2.5pt,
        l sep=6pt,
        fork sep=6pt,
        inner xsep=4pt,
        inner ysep=4pt,
        line width=0.9pt,
        ver/.style={
          rotate=90,
          child anchor=north,
          parent anchor=south,
          anchor=center,
        },
      },
      where level=0{ver, font=\normalsize\bfseries, minimum width=8.5em, minimum height=2.4em, l sep=10pt}{},
      where level=1{ver, coverage category, minimum width=10.8em, minimum height=2.8em, l sep=10pt}{},
      where level=2{coverage function, text width=12.75em, minimum height=2.2em, l sep=6pt}{},
      where level=3{align=left}{},
      [Task Coverage
        [{\mbox{Cross-Modal \textcolor{coverageOrangeFrame}{\textit{Understanding}}}}
          [{Education \& Research}
            [{\coverageindustrytasks{course content structuring\coverageModFlow{\coverageModVideo\coverageModSep\coverageModDocument}{\coverageModText}, research paper summarization\coverageModFlow{\coverageModImage\coverageModSep\coverageModDocument}{\coverageModText}, academic presentation question answering\coverageModFlow{\coverageModVideo\coverageModSep\coverageModDocument}{\coverageModText}, experiment record organization\coverageModFlow{\coverageModImage\coverageModSep\coverageModVideo\coverageModSep\coverageModDocument}{\coverageModText}}}, coverage understanding]
          ]
          [{Business \& Marketing}
            [{\coverageindustrytasks{product demonstration analysis\coverageModFlow{\coverageModVideo\coverageModSep\coverageModDocument}{\coverageModText}, product trial feedback synthesis\coverageModFlow{\coverageModImage\coverageModSep\coverageModAudio\coverageModSep\coverageModDocument}{\coverageModText}, brand material analysis\coverageModFlow{\coverageModImage\coverageModSep\coverageModVideo\coverageModSep\coverageModDocument}{\coverageModText}, sales material summarization\coverageModFlow{\coverageModImage\coverageModSep\coverageModAudio\coverageModSep\coverageModDocument}{\coverageModText}}}, coverage understanding]
          ]
          [{Workplace \& Career}
            [{\coverageindustrytasks{multi-source meeting summarization\coverageModFlow{\coverageModAudio\coverageModSep\coverageModDocument}{\coverageModText}, interview topic analysis\coverageModFlow{\coverageModAudio\coverageModSep\coverageModDocument}{\coverageModText}, job application materials analysis\coverageModFlow{\coverageModVideo\coverageModSep\coverageModDocument}{\coverageModText}}}, coverage understanding]
          ]
          [{Creative Industries}
            [{\coverageindustrytasks{podcast content analysis\coverageModFlow{\coverageModAudio\coverageModSep\coverageModDocument}{\coverageModText}, 3D design analysis\coverageModFlow{\coverageModImage\coverageModSep\coverageModThreeD}{\coverageModText}, game asset inventory\coverageModFlow{\coverageModImage\coverageModSep\coverageModVideo\coverageModSep\coverageModThreeD}{\coverageModText}, exhibit content organization\coverageModFlow{\coverageModImage\coverageModSep\coverageModAudio\coverageModSep\coverageModDocument}{\coverageModText}, event record review\coverageModFlow{\coverageModImage\coverageModSep\coverageModVideo\coverageModSep\coverageModAudio}{\coverageModText}}}, coverage understanding]
          ]
        ]
        [{\mbox{Cross-Modal \textcolor{coverageYellowFrame}{\textit{Generation}}}}
          [{Education \& Science}
            [{\coverageindustrytasks{academic poster generation\coverageModFlow{\coverageModImage\coverageModSep\coverageModDocument}{\coverageModImage}, course video handout\coverageModFlow{\coverageModVideo}{\coverageModDocument}, science explainer video\coverageModFlow{\coverageModDocument}{\coverageModVideo}}}, coverage generation]
          ]
          [{Business \& Marketing}
            [{\coverageindustrytasks{product promotional poster\coverageModFlow{\coverageModImage\coverageModSep\coverageModDocument}{\coverageModImage}, product social-media post\coverageModFlow{\coverageModImage\coverageModSep\coverageModDocument}{\coverageModText\coverageModSep\coverageModImage}, product promotional video\coverageModFlow{\coverageModImage\coverageModSep\coverageModDocument}{\coverageModVideo}}}, coverage generation]
          ]
          [{Media \& Entertainment}
            [{\coverageindustrytasks{video thumbnail design\coverageModFlow{\coverageModVideo}{\coverageModImage}, podcast promotional post\coverageModFlow{\coverageModAudio}{\coverageModText\coverageModSep\coverageModImage}, short-video social post\coverageModFlow{\coverageModVideo}{\coverageModText\coverageModSep\coverageModImage}, game character asset package\coverageModFlow{\coverageModText}{\coverageModImage\coverageModSep\coverageModVideo\coverageModSep\coverageModThreeD}}}, coverage generation]
          ]
          [{Culture \& Events}
            [{\coverageindustrytasks{event invitation design\coverageModFlow{\coverageModText\coverageModSep\coverageModImage}{\coverageModImage}, exhibit audio guide\coverageModFlow{\coverageModImage\coverageModSep\coverageModDocument}{\coverageModAudio}, event teaser video\coverageModFlow{\coverageModText\coverageModSep\coverageModImage}{\coverageModVideo}}}, coverage generation]
          ]
          [{Career Services}
            [{\coverageindustrytasks{job-application introduction video\coverageModFlow{\coverageModDocument}{\coverageModVideo}}}, coverage generation]
          ]
        ]
        [{\mbox{Cross-Modal \textcolor{coverageGreenFrame}{\textit{Reasoning}}}}
          [{Education \& Training}
            [{\coverageindustrytasks{learning path planning\coverageModFlow{\coverageModText\coverageModSep\coverageModDocument}{\coverageModText}}}, coverage reasoning]
          ]
          [{Research \& Science}
            [{\coverageindustrytasks{experimental design optimization\coverageModFlow{\coverageModImage\coverageModSep\coverageModDocument}{\coverageModText}}}, coverage reasoning]
          ]
          [{Software \& Engineering}
            [{\coverageindustrytasks{project debugging\coverageModFlow{\coverageModImage\coverageModSep\coverageModCode}{\coverageModText\coverageModSep\coverageModCode}}}, coverage reasoning]
          ]
          [{Travel \& Tourism}
            [{\coverageindustrytasks{travel route recommendation\coverageModFlow{\coverageModText\coverageModSep\coverageModImage\coverageModSep\coverageModDocument}{\coverageModText}}}, coverage reasoning]
          ]
        ]
        [{\mbox{Cross-Modal \textcolor{coveragePurpleFrame}{\textit{Retrieval}}}}
          [{Technical Support}
            [{\coverageindustrytasks{instruction manual question answering\coverageModFlow{\coverageModImage\coverageModSep\coverageModDocument}{\coverageModText}}}, coverage retrieval]
          ]
          [{Academic Research}
            [{\coverageindustrytasks{related literature retrieval\coverageModFlow{\coverageModDocument}{\coverageModText}}}, coverage retrieval]
          ]
          [{\mbox{Employment \& Recruitment}}
            [{\coverageindustrytasks{job information retrieval\coverageModFlow{\coverageModDocument}{\coverageModText}}}, coverage retrieval]
          ]
          [{News \& Media}
            [{\coverageindustrytasks{related news retrieval\coverageModFlow{\coverageModImage}{\coverageModText}}}, coverage retrieval]
          ]
        ]
      ]
    \end{forest}%
  }
  \caption{Real-world task coverage supported by Omni-IO Skills, organized into cross-modal understanding, generation, reasoning, and retrieval.
  Parenthesized icons indicate representative input-to-output modality flows.}
  \label{fig:task_coverage}
\end{figure}

\paragraph{\textcolor{coverageOrangeFrame}{Cross-Modal Understanding.}}
Understanding Skills convert heterogeneous source material into structured evidence that the host agent can inspect and reuse.
They cover focused media analysis, such as examining a 3D design, and multi-source tasks, such as combining audio with documents for meeting summarization or combining images, video, and 3D assets for a game-asset inventory.
The resulting text can answer the request directly or provide requirements and references for a later generation step.

\paragraph{\textcolor{coverageYellowFrame}{Cross-Modal Generation.}}
Generation Skills support direct transformations, single-asset creation, and coordinated production workflows.
Representative paths include turning a video into a thumbnail, using documents to guide an explainer video, and creating an audio guide from images and exhibit material.
Application-level requests often require several outputs with shared content and style.
A podcast recording can lead to promotional copy and imagery, while a game-character brief can expand into concept art, a showcase video, and a 3D model.

\paragraph{\textcolor{coverageGreenFrame}{Cross-Modal Reasoning.}}
Reasoning Skills use extracted evidence together with user constraints to produce decisions, plans, or revised artifacts.
Figure~\ref{fig:task_coverage} includes learning-path planning, experimental-design optimization, project debugging, and travel-route recommendation.
The debugging flow illustrates that a reasoning task may return both an explanation and modified code.

\paragraph{\textcolor{coveragePurpleFrame}{Cross-Modal Retrieval.}}
Retrieval Skills connect a local task context to information obtained from documents or the web.
They support instruction-manual question answering, related-literature discovery, job-information search, and related-news retrieval.
Returned text can be delivered to the user or passed to another Skill as grounded source material.

\paragraph{Task Composition.}
The task leaves in Figure~\ref{fig:task_coverage} are organized by application domain so that readers can trace a practical request to its modality flow.
The parenthesized icons show representative configurations.
Higher-level Skills can select several leaves, share intermediate assets across them, and assemble the requested deliverables.
Appendix Table~\ref{tab:real_world_task_coverage} provides the corresponding task-to-Skill mappings.

\section{Layered Architecture of Omni-IO Skills}
\label{sec:system_design}

\subsection{Architecture Overview}
\label{sec:architecture_overview}

\begin{figure}[!t]
  \centering
  \includegraphics[width=\linewidth]{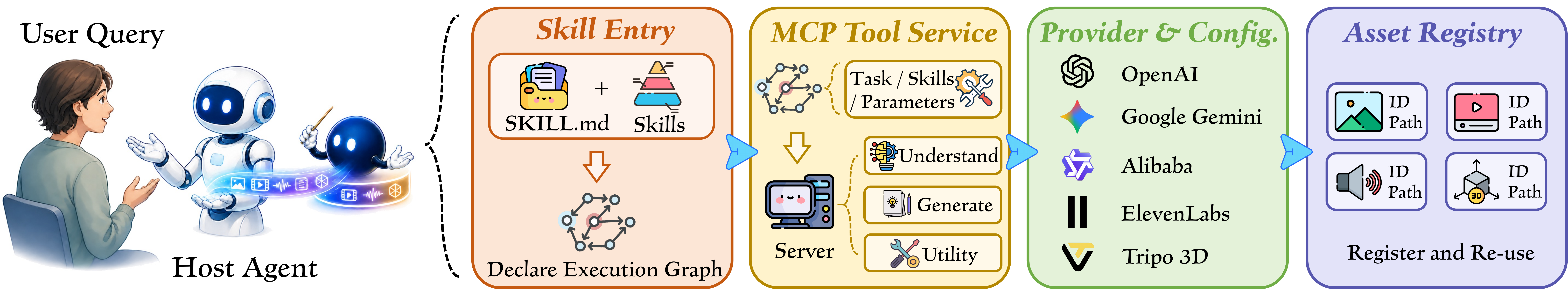}
  \caption{Architecture overview of Omni-IO Skills, which presents the four-layer architecture.}
  \label{fig:architecture_overview}
\end{figure}

As illustrated in Figure~\ref{fig:architecture_overview}, Omni-IO Skills is positioned between the host agent and external multimodal tools.
It adopts a four-layer architecture comprising the \textbf{Skill Entry}, \textbf{MCP Tool Service}, \textbf{Provider and Configuration}, and \textbf{Asset Registry} layers.
These layers separate task knowledge, tool interfaces, service implementations, and persistent outputs, allowing the host agent to plan multimodal tasks without coupling an application workflow to a particular provider or workspace path.

The \textcolor{archentry}{\textbf{Skill Entry}} layer exposes a unified task interface to the host agent and organizes reusable procedural knowledge as Atomic, Expert, and Scenario Skills.
It selects the relevant Skills according to the user request and expands them into executable task specifications.
The \textcolor{archmcp}{\textbf{MCP Tool Service}} layer provides standardized interfaces for multimodal understanding, generation, and utility operations, and maps executable task specifications to the corresponding tool capabilities.
The \textcolor{archprovider}{\textbf{Provider and Configuration}} layer binds those capabilities to concrete providers, models, credentials, default parameters, and fallback policies.
The \textcolor{archregistry}{\textbf{Asset Registry}} layer normalizes and records intermediate and final outputs so that they can be referenced independently of their physical paths and reused by later tasks.

Together, the four layers define a stable interface from procedural knowledge to executable capabilities and persistent artifacts.
The Skill Entry layer expresses the selected workflow as a \textbf{\textit{Declare Execution Graph (DEG)}}, while the lower layers provide the tool, provider, and asset abstractions required to realize its nodes.

\subsection{\texorpdfstring{\textcolor{archentry}{Skill Entry} Layer}{Skill Entry Layer}}
\label{sec:skill_entry}

\begin{figure}[!t]
  \centering
  \includegraphics[width=\linewidth]{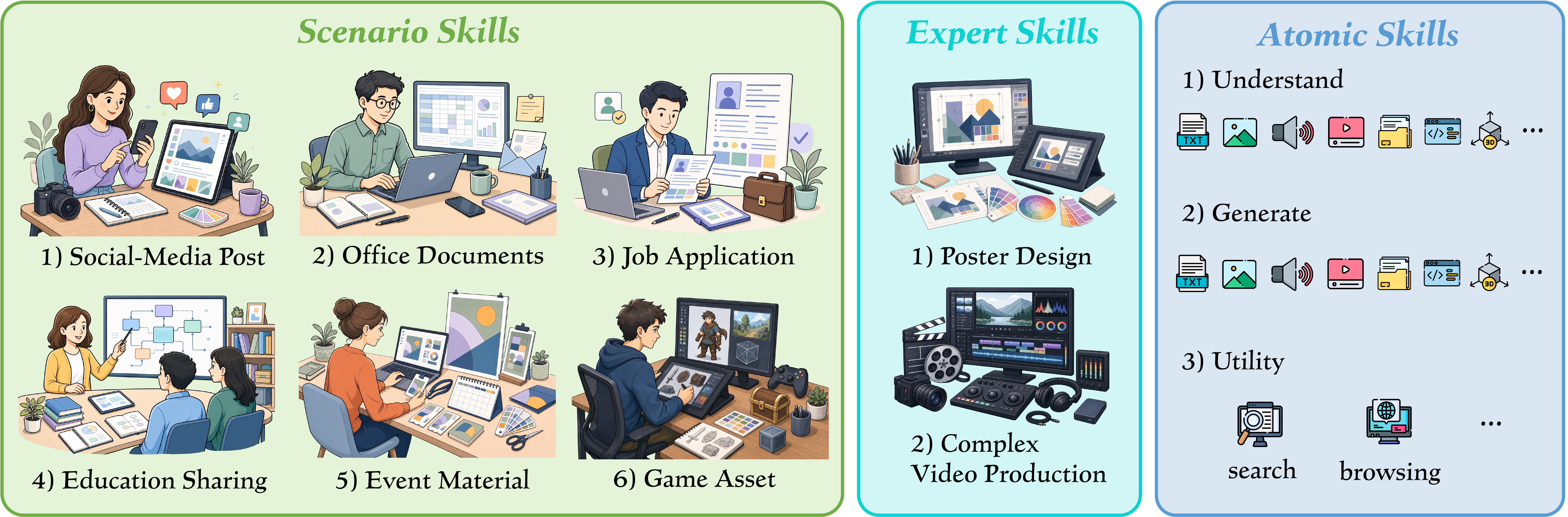}
  \caption{Hierarchical organization of Omni-IO Skills.
  Higher-level Skills organize and invoke lower-level Skills.
  \textcolor{levelscenario}{\textbf{Scenario Skills}} coordinate application-level objectives, \textcolor{levelexpert}{\textbf{Expert Skills}} encapsulate reusable production workflows, and \textcolor{levelatomic}{\textbf{Atomic Skills}} provide executable capability primitives.}
  \label{fig:hierarchical_skills}
\end{figure}

As shown in Figure~\ref{fig:hierarchical_skills}, the Skill Entry layer organizes Skills by task granularity and compositional scope rather than by modality, using three levels: \textcolor{levelatomic}{\textbf{Atomic Skills}}, \textcolor{levelexpert}{\textbf{Expert Skills}}, and \textcolor{levelscenario}{\textbf{Scenario Skills}}.
An Atomic Skill performs a single independently invocable operation.
An Expert Skill targets one concrete final deliverable and organizes multiple atomic operations into a complete workflow.
A Scenario Skill addresses a specific application context, determines the required deliverables from the user's request, and coordinates the appropriate Expert or Atomic Skills.
For example, generating an image is an Atomic operation; producing a poster requires an Expert Skill to coordinate asset generation and layout assembly; and preparing a set of social-media materials may require a Scenario Skill to coordinate both copy and visual assets.
\subsubsection{Declarative Skill Representation and Hierarchical Expansion}
\label{sec:skill_representation}

Skills at all three levels follow a shared declarative representation.
At the logical level, a Skill $s$ can be written as
\begin{equation}
  s = \langle c_s, I_s, P_s, O_s, H_s \rangle,
\end{equation}
where $c_s$ describes its applicability conditions, $I_s$ its required inputs, $P_s$ its execution procedure, $O_s$ its expected outputs, and $H_s$ its relationships to other Skills.
The applicability conditions define the task intent and boundary for which the Skill should be considered.
Inputs and outputs describe artifacts semantically rather than binding them to a particular provider.
The procedure records either a directly executable operation or a workflow that invokes other Skills, and the relationship field identifies the lower-level capabilities available for expansion.
This shared contract allows the host agent to inspect, select, and compose Skills without loading the implementation details of every underlying service.

On this basis, Skill selection begins by identifying the task context and requested deliverables.
A request in a supported application context activates the corresponding Scenario Skill, a request for one professional artifact can select an Expert Skill directly, and a self-contained operation can bypass the upper levels and invoke an Atomic Skill.
After selection, higher-level Skills are recursively expanded until their steps are executable.
Scenario Skills are replaced by the Expert and Atomic Skills required for their selected deliverables, and Expert Skills are replaced by their atomic production and inspection steps.
Table~\ref{tab:implemented_skills} summarizes the implemented Skills and these cross-level invocation and expansion relationships.
Shared inputs and intermediate results are represented once rather than duplicated across deliverables.
The terminal tasks become DEG nodes, including tasks executed natively by the host agent.

\subsubsection{Hierarchical Omni-IO Skills}
\label{sec:hierarchical_skills}

\noindent \textcolor{levelatomic}{\textbf{Atomic Skills.}}
Atomic Skills are the \textbf{\textit{smallest executable units}} in Omni-IO Skills, with each Skill encapsulating a concrete operation for multimodal understanding, content generation, or tool use.
For a request that can be completed in a single step, the host agent directly invokes the corresponding Atomic Skill.
Operations that require external multimodal services are executed through MCP tools, whereas operations such as code or Markdown generation that do not depend on external generation services are performed directly by the host agent.
As stable and reusable execution primitives, Atomic Skills provide the building blocks for Expert and Scenario Skills.

\noindent \textcolor{levelexpert}{\textbf{Expert Skills.}}
Expert Skills target a single final deliverable and encapsulate a complete professional \textit{\textbf{production workflow}}, spanning requirement analysis, task planning, asset generation, final assembly, quality inspection, and localized revision.
When a single Atomic Skill invocation is insufficient to fulfill the user request and the task additionally requires professional decomposition, asset assembly, and final-product inspection, the system selects an appropriate Expert Skill.
It expands its internal workflow into a set of existing Atomic Skills.
Once the atomic tasks have completed, the Expert Skill assembles and inspects the final artifact.
If an issue is detected, only the affected components are regenerated or modified, while validated intermediate assets are reused.
Expert Skills therefore expose professional production capabilities to upper layers without duplicating the underlying Atomic Skills.

\noindent \textcolor{levelscenario}{\textbf{Scenario Skills.}}
Scenario Skills address user tasks situated in explicit application contexts, organizing a set of \textit{\textbf{interrelated deliverables}} around the overall task objective rather than a single modality.
Each Scenario Skill specifies the trigger boundaries, typical inputs, deliverable-selection rules, and expected outputs for its corresponding context, without prescribing a fixed combination of media or execution steps.
The system first determines the task scope from the deliverable types explicitly requested by the user.
When the request is underspecified, it applies context-specific conventions and asks the user for clarification if the task boundary remains ambiguous.
For each deliverable, the Scenario Skill selects an appropriate Expert Skill or directly invokes an Atomic Skill according to its complexity, while coordinating dependencies and asset reuse across different deliverables.

\begin{table}[!t]
  \centering
  \caption{Implemented Omni-IO Skills and their hierarchical composition.}
  \label{tab:implemented_skills}
  \begingroup
  \small
  \setlength{\tabcolsep}{4pt}
  \renewcommand{\arraystretch}{1.18}
  \newcommand{\skillidbadge}[2]{%
    \tikz[baseline=(skillid.base)]{%
      \node[
        inner xsep=0.25pt,
        inner ysep=0.9pt,
        text width=12pt,
        align=center,
        rounded corners=0.8pt,
        fill=#1!14!white,
        text=#1!68!black,
        font=\bfseries\scriptsize
      ] (skillid) {#2};}%
  }
  \newcommand{\skillnamesep}{\hspace{6pt}}
  \newcommand{\skillcallsep}{\hskip 3pt plus 0.6pt\allowbreak}
  \newcommand{\skillarrow}{\ensuremath{\rightarrow}\hspace{5pt}}

  \begin{tabular*}{\linewidth}{@{\hspace{\tabcolsep}\extracolsep{\fill}}>{\raggedright\arraybackslash}p{0.31\linewidth} >{\raggedright\arraybackslash}p{0.31\linewidth} >{\raggedright\arraybackslash}p{0.31\linewidth}}
    \toprule
    \rowcolor{levelatomic!18}
    \multicolumn{3}{c}{\textcolor{levelatomic!68!black}{\textit{\textbf{Atomic Skills}}}} \\
    \midrule
    \rowcolor{levelatomic!4}
    \skillidbadge{levelatomic}{A1}\skillnamesep Image Understanding
      & \skillidbadge{levelatomic}{A2}\skillnamesep Video Understanding
      & \skillidbadge{levelatomic}{A3}\skillnamesep Audio Understanding \\
    \skillidbadge{levelatomic}{A4}\skillnamesep Document Understanding
      & \skillidbadge{levelatomic}{A5}\skillnamesep 3D Understanding
      & \skillidbadge{levelatomic}{A6}\skillnamesep Image Generation \\
    \rowcolor{levelatomic!4}
    \skillidbadge{levelatomic}{A7}\skillnamesep Video Generation
      & \skillidbadge{levelatomic}{A8}\skillnamesep Music Generation
      & \skillidbadge{levelatomic}{A9}\skillnamesep Sound-Effect Generation \\
    \skillidbadge{levelatomic}{A10}\skillnamesep Speech Generation
      & \skillidbadge{levelatomic}{A11}\skillnamesep 3D Generation
      & \skillidbadge{levelatomic}{A12}\skillnamesep PPT Generation \\
    \rowcolor{levelatomic!4}
    \skillidbadge{levelatomic}{A13}\skillnamesep Word Generation
      & \skillidbadge{levelatomic}{A14}\skillnamesep PDF Generation
      & \skillidbadge{levelatomic}{A15}\skillnamesep Excel Generation \\
    \skillidbadge{levelatomic}{A16}\skillnamesep Code Generation
      & \skillidbadge{levelatomic}{A17}\skillnamesep Markdown Generation
      & \skillidbadge{levelatomic}{A18}\skillnamesep Web Search \\
    \rowcolor{levelatomic!4}
    \skillidbadge{levelatomic}{A19}\skillnamesep Web Browsing & & \\
  \end{tabular*}

  \vspace{5pt}
  \begin{tabular*}{\linewidth}{@{\hspace{\tabcolsep}\extracolsep{\fill}}>{\raggedright\arraybackslash}p{0.31\linewidth} >{\raggedright\arraybackslash}p{0.64\linewidth}}
    \toprule
    \rowcolor{levelexpert!18}
    \multicolumn{2}{c}{\textcolor{levelexpert!60!black}{\textit{\textbf{Expert Skills}}}} \\
    \midrule
    \rowcolor{levelexpert!4}
    \skillidbadge{levelexpert}{E1}\skillnamesep Poster Design
      & \skillarrow\skillidbadge{levelatomic}{A1}\skillcallsep
        \skillidbadge{levelatomic}{A6}\skillcallsep
        \skillidbadge{levelatomic}{A16} \\
    \skillidbadge{levelexpert}{E2}\skillnamesep Complex Video Production
      & \skillarrow\skillidbadge{levelatomic}{A2}\skillcallsep
        \skillidbadge{levelatomic}{A3}\skillcallsep
        \skillidbadge{levelatomic}{A6}\skillcallsep
        \skillidbadge{levelatomic}{A7}\skillcallsep
        \skillidbadge{levelatomic}{A8}\skillcallsep
        \skillidbadge{levelatomic}{A9}\skillcallsep
        \skillidbadge{levelatomic}{A10} \\

    \specialrule{0.7pt}{4pt}{\belowrulesep}
    \rowcolor{levelscenario!18}
    \multicolumn{2}{c}{\textcolor{levelscenario!68!black}{\textit{\textbf{Scenario Skills}}}} \\
    \midrule
    \rowcolor{levelscenario!4}
    \skillidbadge{levelscenario}{S1}\skillnamesep Social-Media Post
      & \skillarrow\skillidbadge{levelexpert}{E1}\skillcallsep
        \skillidbadge{levelexpert}{E2}\skillcallsep
        \skillidbadge{levelatomic}{A1}\skillcallsep
        \skillidbadge{levelatomic}{A2}\skillcallsep
        \skillidbadge{levelatomic}{A3}\skillcallsep
        \skillidbadge{levelatomic}{A4}\skillcallsep
        \skillidbadge{levelatomic}{A5}\skillcallsep
        \skillidbadge{levelatomic}{A6}\skillcallsep
        \skillidbadge{levelatomic}{A7}\skillcallsep
        \skillidbadge{levelatomic}{A8}\skillcallsep
        \skillidbadge{levelatomic}{A9}\skillcallsep
        \skillidbadge{levelatomic}{A10} \\
    \skillidbadge{levelscenario}{S2}\skillnamesep Office Documents
      & \skillarrow\skillidbadge{levelatomic}{A3}\skillcallsep
        \skillidbadge{levelatomic}{A4}\skillcallsep
        \skillidbadge{levelatomic}{A12}\skillcallsep
        \skillidbadge{levelatomic}{A13}\skillcallsep
        \skillidbadge{levelatomic}{A14}\skillcallsep
        \skillidbadge{levelatomic}{A15} \\
    \rowcolor{levelscenario!4}
    \skillidbadge{levelscenario}{S3}\skillnamesep Job Application
      & \skillarrow\skillidbadge{levelexpert}{E1}\skillcallsep
        \skillidbadge{levelexpert}{E2}\skillcallsep
        \skillidbadge{levelatomic}{A4}\skillcallsep
        \skillidbadge{levelatomic}{A10}\skillcallsep
        \skillidbadge{levelatomic}{A12}\skillcallsep
        \skillidbadge{levelatomic}{A13} \\
    \skillidbadge{levelscenario}{S4}\skillnamesep Education Sharing
      & \skillarrow\skillidbadge{levelexpert}{E1}\skillcallsep
        \skillidbadge{levelexpert}{E2}\skillcallsep
        \skillidbadge{levelatomic}{A1}\skillcallsep
        \skillidbadge{levelatomic}{A2}\skillcallsep
        \skillidbadge{levelatomic}{A3}\skillcallsep
        \skillidbadge{levelatomic}{A4}\skillcallsep
        \skillidbadge{levelatomic}{A5}\skillcallsep
        \skillidbadge{levelatomic}{A6}\skillcallsep
        \skillidbadge{levelatomic}{A7}\skillcallsep
        \skillidbadge{levelatomic}{A8}\skillcallsep
        \skillidbadge{levelatomic}{A9}\skillcallsep
        \skillidbadge{levelatomic}{A10}\skillcallsep
        \skillidbadge{levelatomic}{A11}\skillcallsep
        \skillidbadge{levelatomic}{A12}\skillcallsep
        \skillidbadge{levelatomic}{A13}\skillcallsep
        \skillidbadge{levelatomic}{A14} \\
    \rowcolor{levelscenario!4}
    \skillidbadge{levelscenario}{S5}\skillnamesep Event Material
      & \skillarrow\skillidbadge{levelexpert}{E1}\skillcallsep
        \skillidbadge{levelexpert}{E2}\skillcallsep
        \skillidbadge{levelatomic}{A6}\skillcallsep
        \skillidbadge{levelatomic}{A7}\skillcallsep
        \skillidbadge{levelatomic}{A8}\skillcallsep
        \skillidbadge{levelatomic}{A9}\skillcallsep
        \skillidbadge{levelatomic}{A10} \\
    \skillidbadge{levelscenario}{S6}\skillnamesep Game Asset
      & \skillarrow\skillidbadge{levelatomic}{A6}\skillcallsep
        \skillidbadge{levelatomic}{A7}\skillcallsep
        \skillidbadge{levelatomic}{A11}\skillcallsep
        \skillidbadge{levelatomic}{A16} \\
    \bottomrule
  \end{tabular*}
  \endgroup
\end{table}

\subsection{\texorpdfstring{\textcolor{archmcp}{MCP Tool Service} Layer}{MCP Tool Service Layer}}
\label{sec:mcp_tool_service}

The MCP Tool Service layer converts the semantic task specifications produced by the Skill Entry layer into standardized executable operations.
It organizes external capabilities into three functional groups: \textbf{understanding} tools that consume multimodal inputs and return structured analyses; \textbf{generation} tools that produce image, video, audio, document, or 3D outputs; and \textbf{utility} tools that provide operations such as search and browsing.
Each externally executed DEG node is submitted through a common contract containing its task type, self-contained prompt, optional parameters, and dependency-resolved asset inputs.
The service uses the task type to select the appropriate tool interface without exposing provider-specific API details to the Skill definition.

The service also defines the boundary between external tool execution and host-native execution.
Operations that depend on specialist multimodal models are invoked through MCP tools, whereas supported operations such as code and Markdown generation can be completed directly by the host agent.
Both routes follow the same DEG dependency semantics and return outputs that can be normalized as system assets.
For external calls, the tool service converts the provider response into a common result containing the output type, local path, description, and effective parameters before forwarding it to the Asset Registry.
Consequently, upper-level Skills can compose heterogeneous capabilities through a stable tool contract even when their underlying APIs and output formats differ.

\subsection{\texorpdfstring{\textcolor{archprovider}{Provider and Configuration} Layer}{Provider and Configuration Layer}}
\label{sec:provider_configuration}

The Provider and Configuration layer separates a tool capability from the service implementation currently used to execute it.
For each externally executable task type, it maintains a binding to the corresponding tool, provider, model, credential reference, default parameters, and fallback policy.
A Skill therefore refers to a semantic capability such as image generation or speech synthesis rather than directly naming a provider-specific endpoint.
The MCP Tool Service consumes this binding when it prepares the concrete invocation.

This separation supports implementation-level substitution without rewriting the procedural knowledge encoded by Skills.
A provider or model can be added, replaced, or reconfigured within the lower layer while the Skill description and DEG structure remain unchanged.
Default parameters provide a consistent baseline for each tool, whereas node-level parameters allow an individual task to specialize the invocation without altering the shared provider configuration.

\subsection{\texorpdfstring{\textcolor{archregistry}{Asset Registry} Layer}{Asset Registry Layer}}
\label{sec:asset_registry}

The Asset Registry provides a shared data abstraction for user-provided, intermediate, and final artifacts.
A registered asset is logically represented as
\begin{equation}
  a = \langle \texttt{asset\_id}, \texttt{type}, \texttt{subtype},
  \texttt{path}, \texttt{description}, \texttt{params},
  \texttt{turn\_id}, \texttt{source\_asset\_id} \rangle.
\end{equation}
The globally unique \texttt{asset\_id} gives upper layers a stable reference that is independent of the physical file path.
The \texttt{type} identifies the broad asset category, while \texttt{subtype} distinguishes finer classes within that category.
The description and effective parameters preserve generation context, and \texttt{turn\_id} associates the artifact with its interaction turn.
When an artifact is revised or derived from an earlier one, \texttt{source\_asset\_id} records the provenance relationship.

The registry applies this representation uniformly to artifacts produced through MCP tools and by the host agent.
Records are persisted in an append-only JSON registry rather than updated in place, allowing multiple versions of a deliverable to remain traceable.
Registry writes are atomic and serialized with a file lock so that parallel nodes or sessions cannot overwrite one another.
Through its lookup and registration interfaces, the layer decouples asset consumers from concrete workspace paths and provides the persistent references required for downstream and cross-turn reuse.

\section{Dependency-Aware Orchestration and Asset Lifecycle}
\label{sec:orchestration}

\subsection{Runtime Workflow}

Upon receiving a user request that involves multimodal understanding or generation, the host agent invokes Omni-IO Skills.
The Skill Entry layer first selects the relevant Skills and expands them into executable tasks.
These tasks are instantiated as nodes in a DEG that explicitly expresses their control and data dependencies.
Before execution, the DEG undergoes structural validation and is then scheduled in successive Waves.
Each externally executed node is mapped to a concrete tool and provider through the MCP Tool Service and Provider and Configuration layers, while natively executed nodes follow the same scheduling protocol.
The resulting assets are registered in the Asset Registry.

\begin{figure}[!t]
  \centering
  \includegraphics[width=\linewidth]{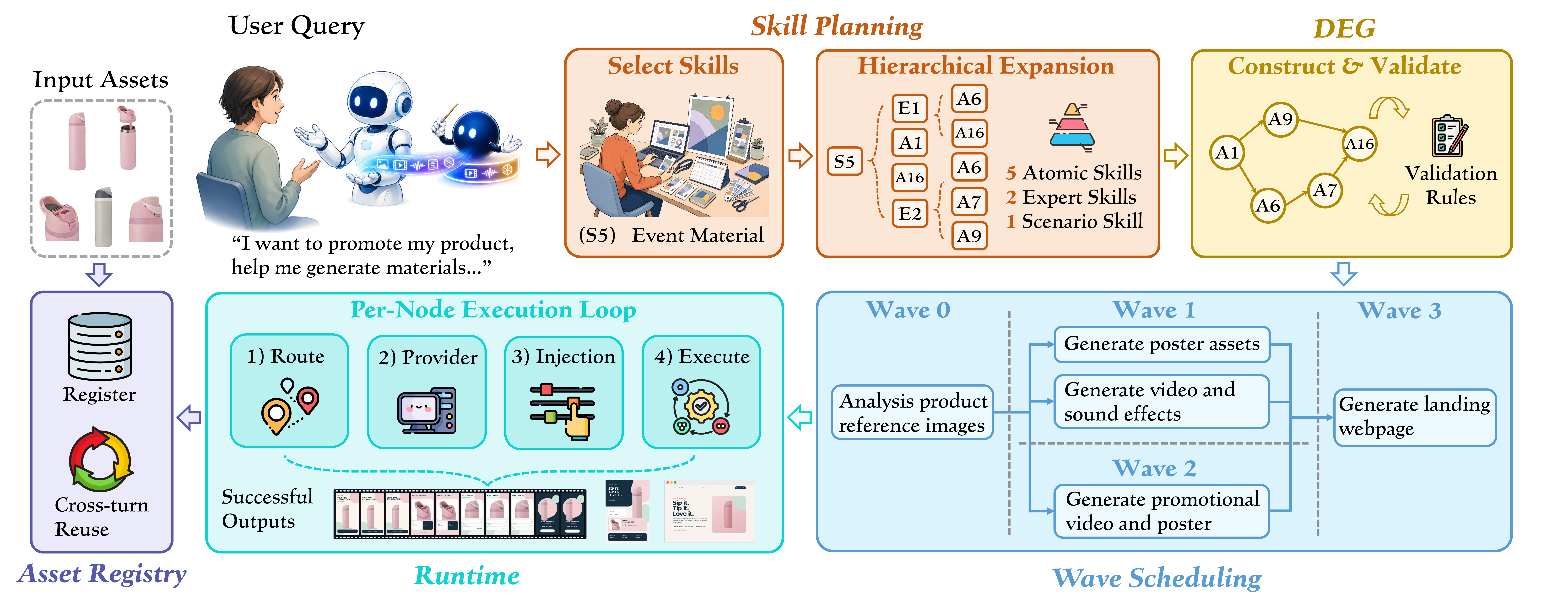}
  \caption{Runtime workflow of Omni-IO Skills, illustrated with a product promotion request.
  The selected S5 Event Material Scenario Skill expands through Expert Skills into five Atomic Skills, whose dependencies form a validated Declare Execution Graph (DEG) and determine successive execution Waves.
  Each ready node follows the same execution loop of tool routing, provider resolution, asset injection, and execution, while successful outputs are registered for downstream use and cross-turn reuse.}
  \label{fig:runtime_workflow}
\end{figure}

The runtime jointly organizes \textbf{\textit{control flow}} and \textbf{\textit{asset flow}}.
Control flow starts from Skill selection and DEG construction, proceeds through validation and dependency-aware scheduling, and ends with the final state of every node.
Asset flow starts from user-provided, historical, and newly generated artifacts.
Each artifact is normalized into a registered asset whose local path can be resolved and passed to downstream nodes.
Figure~\ref{fig:runtime_workflow} summarizes the interaction between these two flows, while Appendix~\ref{app:orchestration_algorithm} provides the corresponding reference procedure.

\subsection{DEG Representation and Construction}

Let $G=(V,E)$ denote a DEG, where each node $v\in V$ is an executable task or an existing-asset source and each directed edge $(u,v)\in E$ indicates that $v$ depends on $u$.
An executable node can be represented as
\begin{equation}
  v = \langle \texttt{id}, \texttt{type}, \texttt{prompt},
  \texttt{params}, \texttt{depends\_on} \rangle.
\end{equation}
The \texttt{id} uniquely identifies the node, \texttt{type} specifies the required capability, \texttt{prompt} provides a self-contained operation description, \texttt{params} specializes the execution, and \texttt{depends\_on} records upstream node or asset references.
An edge acts as a control dependency because the downstream node cannot start before its predecessor completes, and as a data dependency when the predecessor's output is consumed by the downstream operation.

The host agent constructs $G$ from the terminal tasks produced by hierarchical Skill expansion.
Shared intermediate results are represented once and referenced by every consumer.
A historical asset is represented by an \texttt{asset\_source} node containing its \texttt{asset\_ref}; it is considered completed after the reference is resolved and issues no generation call.

\subsection{Graph Validation, Wave Scheduling, and Failure Handling}

Before execution, the DEG is validated to ensure that every dependency reference resolves to a node or registered asset and that the induced graph is acyclic.
A graph that fails either check is treated as a planning error and is reconstructed rather than submitted in an invalid form.

For a valid DEG, pending nodes whose predecessors have all completed form the next execution Wave.
Nodes in the same Wave are mutually independent and are dispatched concurrently.
Nodes with no unresolved predecessors enter the earliest Wave, and every remaining node enters the first later Wave in which all of its immediate predecessors have completed.
The schedule is determined by declared dependencies rather than modality: image and audio generation can run concurrently when neither consumes the other, whereas an image-conditioned video node waits for its reference image.

When a node fails, its pending descendants are cancelled because their required inputs are unavailable, while nodes on independent branches continue to execute.
The system records the final state of every node and does not roll back outputs produced by completed nodes.

\subsection{Runtime Tool Routing and Parameter Injection}
\label{sec:runtime_tool_routing}

For each executable node, the MCP Tool Service selects a tool according to the node's task type.
The Provider and Configuration layer then resolves the corresponding provider, model, credentials, default parameters, and fallback policy.
At invocation time, the effective parameters combine provider defaults, node-level overrides, and inputs injected from predecessor outputs.

Once an upstream node completes, the system resolves the local paths of its outputs and passes them to the downstream tool call according to the task relationship.
An image-to-video dependency supplies the image as the initial visual condition, whereas an image-to-3D dependency supplies it as the reference image.
Because these inputs are injected from the graph, downstream prompts do not need to embed paths.
Code, Markdown, and other supported tasks executed natively by the host agent bypass MCP generation services but retain the same input-resolution, status, and output-registration contract.

\subsection{Asset Propagation and Cross-Turn Reuse}
\label{sec:runtime_asset_flow}

Once a node completes successfully, its output is written to the workspace and registered in the Asset Registry.
MCP-tool outputs are registered automatically when their results are returned, whereas files produced natively by the host agent are registered explicitly after they are written.

Within one turn, downstream nodes resolve upstream outputs through their asset identifiers and consume the corresponding local files.
Across turns, a user can refer to a historical asset through \texttt{asset\_ref}; the referenced record is materialized as an already-completed source node in the new DEG and becomes immediately available to its descendants.
A revision creates a new record that may retain a reference to its source instead of updating the existing record in place.

Registration occurs only after the corresponding output has been written successfully.
Registry writes are atomic and serialized with a file lock so that concurrent nodes or sessions do not overwrite one another, and every completed registry entry resolves to a concrete artifact.

Consider a request to promote a product from a set of product images by producing a poster, a promotional video with sound effects, and a landing web page.
The Event Material Scenario Skill expands the request through Poster Design and Complex Video Production into the Atomic Skills required for image understanding, image generation, video generation, sound-effect generation, and code generation.
As shown in Table~\ref{tab:deg_example}, the DEG schedules these tasks in four Waves: product-reference analysis first, two parallel asset-generation tasks second, promotional-content production third, and landing-page generation last.
If the user later requests a new presentation style for the landing page, the registered product analysis and promotional assets are materialized as completed \texttt{asset\_source} nodes, and only the landing-page task is executed again.

\begin{table}[!t]
  \centering
  \caption{Illustrative four-Wave DEG execution for the product-promotion case in Figure~\ref{fig:runtime_workflow}.}
  \label{tab:deg_example}
  \small
  \setlength{\tabcolsep}{1.2pt}
  \renewcommand{\arraystretch}{1.18}
  \begin{tabular*}{\linewidth}{@{\extracolsep{\fill}}>{\raggedright\arraybackslash}p{0.09\linewidth} >{\raggedright\arraybackslash}p{0.35\linewidth} >{\raggedright\arraybackslash}p{0.55\linewidth}@{}}
    \toprule
    \textbf{Stage} & \textbf{Node(s)} & \textbf{Dependency and output} \\
    \midrule
    Wave 0 & Analyze product reference images & Extracts product appearance and visual constraints. \\
    Wave 1 & Generate poster assets & Generates poster-ready visuals from the product analysis. \\
           & Generate video and sound effects & Generates video and sound-effect assets in parallel. \\
    Wave 2 & Generate promotional video and poster & Combines assets into the promotional video and poster. \\
    Wave 3 & Generate landing web page & Assembles a landing page from the promotional outputs. \\
    \bottomrule
  \end{tabular*}
\end{table}

\section{Experiments}

\subsection{Experimental Setup}
\noindent \textbf{Baselines.}
We evaluate two general-purpose agents, GPT-5.6 Sol~\cite{openai2026gpt56} and Claude Sonnet 5~\cite{anthropic2026sonnet5}, under two configurations for each agent: \textit{Base Agent} and \textit{Agent + Omni-IO Skills}.
The host agent runs in its default environment, retaining the built-in Skills and tools provided by the system while excluding any additionally installed third-party extensions or task-specific customizations.
\textit{Agent + Omni-IO Skills} further loads Omni-IO Skills and its associated MCP tool services into the same environment, while keeping all other built-in Skills, tool configurations, prompts, and execution budgets unchanged.
Because the default capabilities of the two agents may differ, we focus on the within-agent gains introduced by Omni-IO Skills and do not interpret cross-agent score differences as a ranking of the models' capabilities.

\noindent \textbf{Benchmark and evaluation metrics.}
We construct UniM-90 by selecting a fixed subset of 90 instances from UniM~\cite{li2026unim}, covering text, image, audio, video, document, code, and 3D modalities, together with their representative interleaved combinations.
The subset is selected independently of the native modality capabilities of any evaluated agent; the detailed selection protocol is provided in Appendix~\ref{app:experiment:unim90}.
For evaluation, we adopt the UniM Evaluation Suite and report the input-support rate $\tau$, together with the absolute and relative variants of Semantic--Quality Coupled Score (SQCS), Interleaved Coherence Score (ICS), Strict Structure Score (StS), and Lenient Structure Score (LeS).
Here, $\tau$ denotes the proportion of instances for which an agent can fully receive and process all input modalities; the absolute score $\mathcal{X}^{\mathrm{abs}}$ measures performance over the subset of supported instances, whereas the relative score $\mathcal{X}^{\mathrm{rel}}=\tau\mathcal{X}^{\mathrm{abs}}$ further reflects performance over the complete test set.
SQCS jointly evaluates semantic correctness and generation quality; StS and LeS measure strict output-structure consistency and modality-level coverage, respectively; and ICS assesses the overall coherence of interleaved multimodal responses.
For instances whose inputs can be processed but whose required output modalities cannot be generated or whose target outputs cannot be completed, the resulting failures remain included in metric computation.

\subsection{Main Results}
\begin{table}[!t]
  \centering
  \caption{Main results on UniM-90.
  We report the input-support rate $\tau$ together with the absolute and relative variants of Semantic--Quality Coupled Score (SQCS), Interleaved Coherence Score (ICS), Strict Structure Score (StS), and Lenient Structure Score (LeS).
  Higher is better for all metrics.}
  \label{tab:main_results}
  \fontsize{9}{9}\selectfont
  \setlength{\tabcolsep}{2.5mm}
  \renewcommand{\arraystretch}{1.4}
  \begin{tabular}{l c cccc cccc}
    \toprule
    \multirow{2}{*}{\textbf{Base Agent}} & \multirow{2}{*}{$\boldsymbol{\tau}$}
    & \multicolumn{4}{c}{\textbf{Absolute}}
    & \multicolumn{4}{c}{\textbf{Relative}} \\
    \cmidrule(lr){3-6}
    \cmidrule(lr){7-10}
    & & \textbf{SQCS} & \textbf{ICS} & \textbf{StS} & \textbf{LeS} & \textbf{SQCS} & \textbf{ICS} & \textbf{StS} & \textbf{LeS} \\
    \midrule
    GPT-5.6 Sol & 40\% & 67.49 & 86.53 & 47.84 & 72.22 & 26.99 & 34.61 & 19.14 & 28.89 \\
    \quad + \textbf{Omni-IO Skills} & 100\% & 74.94 & 93.98 & 100.00 & 100.00 & 74.94 & 93.98 & 100.00 & 100.00 \\
    \midrule
    Claude Sonnet 5 & 38.89\% & 71.53 & 82.29 & 52.21 & 68.57 & 27.82 & 32.00 & 20.30 & 26.67 \\
    \quad + \textbf{Omni-IO Skills} & 100\% & 77.78 & 83.28 & 99.78 & 100.00 & 77.78 & 83.28 & 99.78 & 100.00 \\
    \bottomrule
  \end{tabular}%
\end{table}

As shown in Table~\ref{tab:main_results}, Omni-IO Skills yields substantial and consistent improvements for both base agents. First, the input-support rates of GPT-5.6 Sol and Claude Sonnet 5 increase from 40.00\% and 38.89\%, respectively, to 100\% in both cases. This result shows that Omni-IO Skills effectively addresses the limitations of base agents in handling heterogeneous multimodal inputs and outputs, including audio, video, documents, and 3D content, thereby enabling them to cover all tasks in UniM-90. In terms of the relative metrics, SQCS increases from 26.99 and 27.82 to 74.94 and 77.78, while ICS increases from 34.61 and 32.00 to 93.98 and 83.28, respectively, demonstrating substantial improvements in semantic quality and interleaved multimodal coherence over the full test set. For the absolute metrics, SQCS increases from 67.49 and 71.53 to 74.94 and 77.78, respectively, indicating that Omni-IO Skills not only broadens modality support but also improves the semantic correctness and generation quality of task outputs. The gains are particularly pronounced for the structure metrics: GPT-5.6 Sol achieves 100\% on both StS and LeS, while Claude Sonnet 5 reaches 99.78\% and 100\%, respectively, validating the effectiveness of Omni-IO Skills in multimodal tool invocation, asset organization, and output-structure control.

\subsection{Case Studies}
\begin{figure}[!t]
  \centering
  \includegraphics[width=\linewidth]{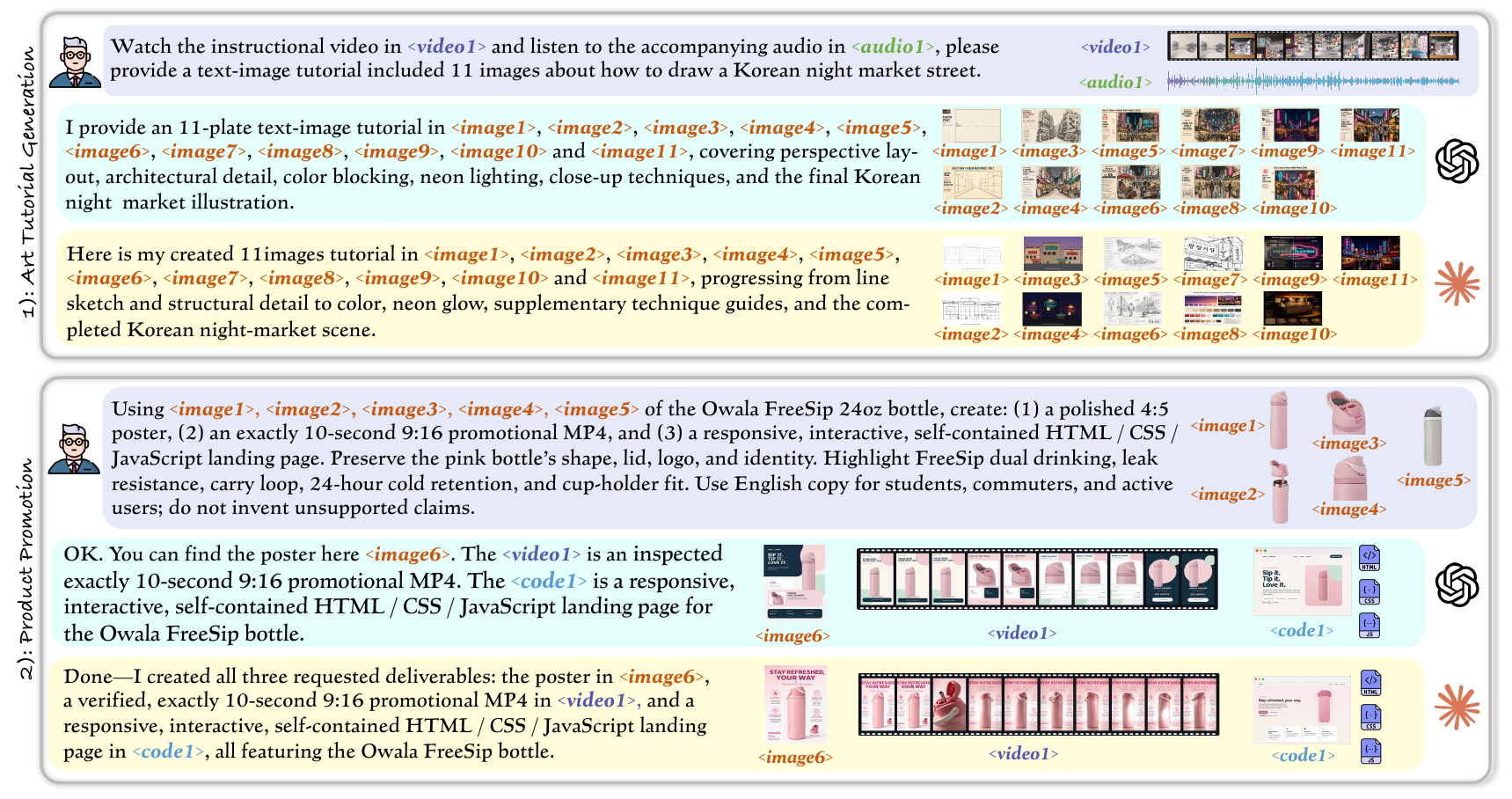}
  \caption{Qualitative case studies of Omni-IO Skills with GPT-5.6 Sol and Claude Sonnet 5.}
  \vspace{-2pt}
  \label{fig:case_studies}
\end{figure}

As shown in Figure~\ref{fig:case_studies}, we qualitatively analyze the end-to-end performance of Omni-IO Skills with GPT-5.6 Sol and Claude Sonnet 5 through two cases: art tutorial generation and product promotion. In the art tutorial generation case, Omni-IO Skills invokes the S4 Education Sharing Skill to plan the tutorial and coordinate the workflow, the A2 Video Understanding Skill to extract the drawing steps from the video, the A3 Audio Understanding Skill to recover the accompanying explanations and procedural order, and the A6 Image Generation Skill to produce the 11-image tutorial. In the product promotion case, Omni-IO Skills invokes the A1 Image Understanding Skill to extract product attributes from the images, the E1 Poster Design Skill to design the poster, the E2 Complex Video Production Skill to produce the promotional video, and the A16 Code Generation Skill to build the landing page, thereby generating the requested deliverables. Both GPT-5.6 Sol and Claude Sonnet 5 produce all requested outputs while preserving consistency across tutorial steps, product identity, and deliverables, demonstrating the stable performance of Omni-IO Skills in multimodal understanding, generation, and complex task orchestration.

\section{Conclusion}

Omni-IO Skills presents a plug-and-play Agent Harness that makes general-purpose agents omni-native without altering their reasoning core.
The system organizes multimodal capabilities as hierarchical Skills, maps complex requests into dependency-aware DEGs, routes tasks across replaceable specialist backends, and preserves generated artifacts through a persistent Asset Registry.
On UniM-90, the harness raises the input-support rates of GPT-5.6 Sol and Claude Sonnet 5 to 100\%, increases relative Semantic--Quality Coupled Score by 47.95 and 49.96 points, and reaches Strict Structure Scores of 100.00 and 99.78.
These results show that harness-level composition provides a practical route to Omni capability while keeping procedures, providers, and assets independently extensible.
Omni-IO Skills therefore offers an application-oriented foundation for agents that coordinate heterogeneous media and reusable outputs across multi-turn production workflows.

\bibliography{colm2024_conference}
\bibliographystyle{unsrt}

\newpage
\appendix
\etocdepthtag.toc{appendixmatter}
\begingroup
  \setlength{\cftbeforesecskip}{2pt}
  \setlength{\cftbeforesubsecskip}{0pt}
  \etocsettagdepth{mainmatter}{none}
  \etocsettagdepth{appendixmatter}{subsection}
  \etocsetnexttocdepth{subsection}
  \etocsettocstyle{\section*{Appendix Contents}}{}
\tableofcontents
\endgroup
\newpage
\section{Complete Catalog of Omni-IO Skills}
{%
  \fontsize{8.5}{8.5}\selectfont
  \setlength{\tabcolsep}{2mm}
  \renewcommand{\arraystretch}{1.25}
  \hypertarget{table.\number\numexpr\value{table}+1\relax}{}
  \begin{xltabular}{\linewidth}{@{}l@{\hspace{2mm}}l@{\hspace{2mm}}>{\arraybackslash}X@{}}
    \caption{Complete catalog of Omni-IO Skills across the Atomic, Expert, and Scenario Skills.}
    \label{tab:skill_catalog}\\
    \toprule
    \textbf{ID} & \textbf{Skill} & \textbf{Description} \\
    \midrule
    \endfirsthead

    \multicolumn{3}{c}{\tablename\ \thetable\ (continued)} \\
    \toprule
    \textbf{ID} & \textbf{Skill} & \textbf{Description} \\
    \midrule
    \endhead

    \midrule
    \multicolumn{3}{r}{Continued on next page} \\
    \endfoot

    \bottomrule
    \endlastfoot

    \multicolumn{3}{c}{\cellcolor{catgray}\textit{\textbf{Atomic Skills}}} \\
    \midrule
    A1 & Image Understanding & Analyzes image content, text, objects, visual style, composition, and atmosphere. \\
    A2 & Video Understanding & Summarizes video content and analyzes timelines, speech, key frames, and audiovisual style. \\
    A3 & Audio Understanding & Transcribes speech and analyzes music, sound effects, speakers, timing, and acoustic scenes. \\
    A4 & Document Understanding & Extracts and summarizes text, structure, tables, and slide content from PDF, Word, PowerPoint, and Excel files. \\
    A5 & 3D Understanding & Inspects 3D geometry, dimensions, materials, topology, semantic category, and rendered appearance. \\
    \midrule
    A6 & Image Generation & Generates images from text prompts with configurable aspect ratios and output formats. \\
    A7 & Video Generation & Generates text- or image-conditioned videos with configurable duration, resolution, and aspect ratio. \\
    A8 & Music Generation & Generates instrumental or vocal music from descriptions of style, mood, instrumentation, and tempo. \\
    A9 & Sound-Effect Generation & Generates environmental sounds and effects with controllable duration, source, intensity, and spatial character. \\
    A10 & Speech Generation & Synthesizes multilingual speech with configurable voice, language, and speaking speed. \\
    A11 & 3D Generation & Generates GLB models from text prompts or reference images for downstream visualization and production. \\
    A12 & PPT Generation & Produces structured PowerPoint presentations from agent-planned slide titles and key points. \\
    A13 & Word Generation & Produces structured Word documents from agent-planned headings and paragraphs. \\
    A14 & PDF Generation & Produces structured PDF documents from agent-planned sections and body text. \\
    A15 & Excel Generation & Produces styled, multi-sheet Excel workbooks from structured headers and data rows. \\
    A16 & Code Generation & Creates and registers code or web artifacts, supports versioned iteration, and integrates multiple upstream assets. \\
    A17 & Markdown Generation & Creates and registers structured Markdown documents, supports versioned edits, and integrates multiple upstream assets. \\
    \midrule
    A18 & Web Search & Retrieves up-to-date web information and returns structured results with source links. \\
    A19 & Web Browsing & Navigates webpages, interacts with page elements, and extracts content through a local browser. \\

    \midrule
    \multicolumn{3}{c}{\cellcolor{catgray}\textit{\textbf{Expert Skills}}} \\
    \midrule
    E1 & Poster Design & Creates and reviews polished posters, invitations, promotional graphics, and social covers by combining generated or supplied visuals with exact typography and deterministic layout. \\
    E2 & Complex Video Production & Produces and reviews complete multi-scene videos by coordinating scripts, storyboards, visual clips, narration, music, subtitles, branding, and final assembly. \\

    \midrule
    \multicolumn{3}{c}{\cellcolor{catgray}\textit{\textbf{Scenario Skills}}} \\
    \midrule
    S1 & Social-Media Post & Orchestrates platform-aware copy with suitable image, video, or audio assets for social publishing. \\
    S2 & Office Documents & Converts text, recordings, and source documents into structured presentations, reports, minutes, or spreadsheets. \\
    S3 & Job Application & Creates resumes, cover letters, self-introduction presentations, speech, or video for a target role. \\
    S4 & Education Sharing & Produces teaching and explanatory materials such as slides, documents, illustrations, narration, and videos. \\
    S5 & Event Material & Creates invitations, posters, atmosphere music, and promotional or retrospective videos for events. \\
    S6 & Game Asset & Orchestrates concept art, 3D models, showcase videos, and optional web presentations for game assets. \\
  \end{xltabular}
}

\section{Representative Real-World Task Coverage}
{%
  \newcommand{\modimage}{\raisebox{-0.18\height}{\includegraphics[height=1.1em]{logo/image.png}}}
  \newcommand{\modvideo}{\raisebox{-0.18\height}{\includegraphics[height=1.1em]{logo/video.png}}}
  \newcommand{\modaudio}{\raisebox{-0.18\height}{\includegraphics[height=1.1em]{logo/audio.png}}}
  \newcommand{\moddocument}{\raisebox{-0.18\height}{\includegraphics[height=1.1em]{logo/document.png}}}
  \newcommand{\modthreeD}{\raisebox{-0.18\height}{\includegraphics[height=1.1em]{logo/3d.png}}}
  \newcommand{\modtext}{\raisebox{-0.18\height}{\includegraphics[height=1em]{logo/text.png}}}
  \newcommand{\modcode}{\raisebox{-0.18\height}{\includegraphics[height=1.1em]{logo/code.png}}}
  \newcommand{\modsep}{\hspace{4pt}}
  \setlength{\abovecaptionskip}{0pt}
  \setlength{\belowcaptionskip}{-4pt}
  \fontsize{8.5}{8.5}\selectfont
  \setlength{\tabcolsep}{3mm}
  \renewcommand{\arraystretch}{1.25}
  \newlength{\tablethreeskillidwidth} \settowidth{\tablethreeskillidwidth}{A1, A4, A18, A19} \newlength{\tablethreeinputwidth}
  \setlength{\tablethreeinputwidth}{24mm}
  \newlength{\tablethreeoutputwidth}
  \setlength{\tablethreeoutputwidth}{21mm}
  \hypertarget{table.\number\numexpr\value{table}+1\relax}{}
  \begin{xltabular}{\linewidth}{@{}>{\arraybackslash}X@{\hspace{5mm}}>{\raggedright\arraybackslash}p{\tablethreeskillidwidth}@{\hspace{5mm}}>{\centering\arraybackslash}p{\tablethreeinputwidth}@{\hspace{5mm}}>{\centering\arraybackslash}p{\tablethreeoutputwidth}@{}}
    \caption[Representative real-world task coverage of Omni-IO Skills.]{Representative real-world task coverage of Omni-IO Skills.
    Both Skill IDs and modality combinations denote representative configurations rather than exhaustive implementations.
    Icons denote \modtext\,text, \modimage\,image, \modaudio\,audio, \modvideo\,video, \moddocument\,document, \modcode\,code, and \modthreeD\,3D.}
    \label{tab:real_world_task_coverage}\\
    \toprule
    \textbf{Task} & \textbf{Skill ID(s)} & \makecell[c]{\textbf{Input} \\ \textbf{Modality(s)}} & \makecell[c]{\textbf{Output} \\ \textbf{Modality(s)}} \\
    \midrule
    \endfirsthead

    \multicolumn{4}{c}{\tablename\ \thetable\ (continued)} \\
    \toprule
    \textbf{Task} & \textbf{Skill ID(s)} & \makecell[c]{\textbf{Input} \\ \textbf{Modality(s)}} & \makecell[c]{\textbf{Output} \\ \textbf{Modality(s)}} \\
    \midrule
    \endhead

    \midrule
    \multicolumn{4}{r}{Continued on next page} \\
    \endfoot

    \bottomrule
    \endlastfoot

    \multicolumn{4}{c}{\cellcolor{catgray}\textit{\textbf{Cross-Modal Understanding}}} \\
    \midrule
    Multi-Source Meeting Summarization & A3, A4 & \modaudio\modsep\moddocument & \modtext \\
    Interview Topic Analysis & A3, A4 & \modaudio\modsep\moddocument & \modtext \\
    Course Content Structuring & A2, A4 & \modvideo\modsep\moddocument & \modtext \\
    Research Paper Summarization & A1, A4 & \modimage\modsep\moddocument & \modtext \\
    Academic Presentation Question Answering & A2, A4 & \modvideo\modsep\moddocument & \modtext \\
    Product Demonstration Analysis & A2, A4 & \modvideo\modsep\moddocument & \modtext \\
    Podcast Content Analysis & A3, A4 & \modaudio\modsep\moddocument & \modtext \\
    Product Trial Feedback Synthesis & A1, A3, A4 & \modimage\modsep\modaudio\modsep\moddocument & \modtext \\
    Brand Material Analysis & A1, A2, A4 & \modimage\modsep\modvideo\modsep\moddocument & \modtext \\
    Job Application Materials Analysis & A2, A4 & \modvideo\modsep\moddocument & \modtext \\
    3D Design Analysis & A1, A5 & \modimage\modsep\modthreeD & \modtext \\
    Game Asset Inventory & A1, A2, A5 & \modimage\modsep\modvideo\modsep\modthreeD & \modtext \\
    Exhibit Content Organization & A1, A3, A4 & \modimage\modsep\modaudio\modsep\moddocument & \modtext \\
    Sales Material Summarization & A1, A3, A4 & \modimage\modsep\modaudio\modsep\moddocument & \modtext \\
    Event Record Review & A1, A2, A3 & \modimage\modsep\modvideo\modsep\modaudio & \modtext \\
    Experiment Record Organization & A1, A2, A4 & \modimage\modsep\modvideo\modsep\moddocument & \modtext \\
    \midrule
    \multicolumn{4}{c}{\cellcolor{catgray}\textit{\textbf{Cross-Modal Generation}}} \\
    \midrule
    Academic Poster Generation & E1, A1, A4 & \modimage\modsep\moddocument & \modimage \\
    Video Thumbnail Design & E1, A2 & \modvideo & \modimage \\
    Podcast Promotional Post & S1, E1, A3 & \modaudio & \modtext\modsep\modimage \\
    Product Promotional Poster & E1, A1, A4 & \modimage\modsep\moddocument & \modimage \\
    Event Invitation Design & S5, E1, A1 & \modtext\modsep\modimage & \modimage \\
    Product Social-Media Post & S1, A1, A4, A6 & \modimage\modsep\moddocument & \modtext\modsep\modimage \\
    Short-Video Social Post & S1, E1, A2 & \modvideo & \modtext\modsep\modimage \\
    Course Video Handout & S4, A2, A13 & \modvideo & \moddocument \\
    Exhibit Audio Guide & A1, A4, A10 & \modimage\modsep\moddocument & \modaudio \\
    Product Promotional Video & S1, E2, A1, A4 & \modimage\modsep\moddocument & \modvideo \\
    Science Explainer Video & S4, E2, A4 & \moddocument & \modvideo \\
    Event Teaser Video & S5, E2, A1 & \modtext\modsep\modimage & \modvideo \\
    Job-Application Introduction Video & S3, E2, A4 & \moddocument & \modvideo \\
    Game Character Asset Package & S6, A6, A7, A11 & \modtext & \modimage\modsep\modvideo\modsep\modthreeD \\
    \midrule
    \multicolumn{4}{c}{\cellcolor{catgray}\textit{\textbf{Cross-Modal Reasoning}}} \\
    \midrule
    Learning Path Planning & A4, A18, A19 & \modtext\modsep\moddocument & \modtext \\
    Experimental Design Optimization & A1, A4 & \modimage\modsep\moddocument & \modtext \\
    Project Debugging & A1, A16 & \modimage\modsep\modcode & \modtext\modsep\modcode \\
    Travel Route Recommendation & A1, A4, A18, A19 & \modtext\modsep\modimage\modsep\moddocument & \modtext \\
    \midrule
    \multicolumn{4}{c}{\cellcolor{catgray}\textit{\textbf{Cross-Modal Retrieval}}} \\
    \midrule
    Instruction Manual Question Answering & A1, A4 & \modimage\modsep\moddocument & \modtext \\
    Related Literature Retrieval & A4, A18, A19 & \moddocument & \modtext \\
    Job Information Retrieval & A4, A18, A19 & \moddocument & \modtext \\
    Related News Retrieval & A1, A18, A19 & \modimage & \modtext \\
  \end{xltabular}
}

\section{MCP Tools and Provider Bindings}
\label{app:mcp}

\subsection{Tool Inventory}
\label{app:mcp:inventory}

Table~\ref{tab:tool_inventory} lists the tools of the MCP Tool Service layer, grouped into understanding, generation, utility, and registry management.
Capabilities that require an external service are exposed as MCP tools.
Image understanding uses the host agent's native vision, while code and Markdown generation use the host agent's native file writing.
Generation tools register their outputs before returning; the natively executed \texttt{code} and \texttt{markdown} task types require an explicit \texttt{register\_asset} call.
Document generation tools receive a planned \texttt{structure} and content from the host agent and render them into the requested document files.

\begin{table}[!t]
  \centering
  \caption{Tools exposed by the MCP Tool Service layer.
  \emph{Registers} indicates whether the tool writes its output to the Asset Registry before returning.
  Image understanding, code generation, and Markdown generation are executed natively by the host agent.}
  \label{tab:tool_inventory}
  \fontsize{8.5}{9.5}\selectfont
  \setlength{\tabcolsep}{2mm}
  \renewcommand{\arraystretch}{1.2}
  \begin{tabularx}{\linewidth}{@{}l l >{\raggedright\arraybackslash}X c@{}}
    \toprule
    \textbf{Tool} & \textbf{Modality} & \textbf{Key parameters and returned content} & \textbf{Registers} \\
    \midrule
    \multicolumn{4}{c}{\cellcolor{catgray}\textit{\textbf{Understanding}}} \\
    \midrule
    \texttt{understand\_video} & video & \texttt{video\_url}; returns a summary description and a segmented timeline & \xmark \\
    \texttt{understand\_audio} & audio & \texttt{audio\_url}, \texttt{subtype}; returns a transcript for speech and a style analysis for music and sound effects & \xmark \\
    \texttt{understand\_document} & document & \texttt{document\_url}; returns a summary and the extracted body text in Markdown & \xmark \\
    \texttt{understand\_3d} & 3D & \texttt{file\_url}; returns mesh metadata, rendered projection paths, and the renderer actually used & \xmark \\
    \midrule
    \multicolumn{4}{c}{\cellcolor{catgray}\textit{\textbf{Generation}}} \\
    \midrule
    \texttt{generate\_image} & image & \texttt{prompt}, \texttt{aspect\_ratio}, \texttt{quality} & \cmark \\
    \texttt{generate\_video} & video & \texttt{prompt}, \texttt{duration\_seconds}, \texttt{aspect\_ratio}, \texttt{first\_frame\_url} & \cmark \\
    \texttt{generate\_music} & audio & \texttt{prompt}, \texttt{duration\_seconds}, \texttt{instrumental} & \cmark \\
    \texttt{generate\_sfx} & audio & \texttt{prompt}, \texttt{duration\_seconds} & \cmark \\
    \texttt{generate\_speech} & audio & \texttt{text}, \texttt{voice}, \texttt{language}, \texttt{speed} & \cmark \\
    \texttt{generate\_3d} & 3D & \texttt{prompt}, \texttt{reference\_image\_url}; always emits GLB & \cmark \\
    \texttt{generate\_ppt} & document & \texttt{structure} (title and per-slide key points), \texttt{style} & \cmark \\
    \texttt{generate\_word} & document & \texttt{structure} (title, headings, paragraphs), \texttt{style} & \cmark \\
    \texttt{generate\_pdf} & document & \texttt{structure} (title, headings, paragraphs), \texttt{style} & \cmark \\
    \texttt{generate\_excel} & document & \texttt{structure} (per-sheet headers and rows), \texttt{style} & \cmark \\
    \midrule
    \multicolumn{4}{c}{\cellcolor{catgray}\textit{\textbf{Utility}}} \\
    \midrule
    \texttt{search} & text & \texttt{query}, \texttt{count}; returns titles, links, and snippets & \xmark \\
    \texttt{browse} & text & \texttt{url}, \texttt{max\_chars}; returns extracted page text with a truncation flag & \xmark \\
    \texttt{check\_config} & --- & returns ready, missing, and always-free capabilities together with the active providers & \xmark \\
    \midrule
    \multicolumn{4}{c}{\cellcolor{catgray}\textit{\textbf{Registry management}}} \\
    \texttt{set\_turn} & --- & \texttt{turn\_id}; stamps every asset registered in the current interaction turn & \xmark \\
    \texttt{register\_asset} & any & \texttt{asset\_type}, \texttt{subtype}, \texttt{path}, \texttt{description}, \texttt{params}, \texttt{source\_asset\_id}; used for natively produced files & \cmark \\
    \texttt{get\_asset} & any & \texttt{asset\_id}; returns the full record, including the local path & \xmark \\
    \texttt{list\_assets} & any & returns all registered assets in reverse completion order & \xmark \\
    \bottomrule
  \end{tabularx}
\end{table}

\subsection{Provider Bindings and Fallback}
\label{app:mcp:providers}

Table~\ref{tab:provider_bindings} lists the provider, model, and API key bound to each capability in the configuration used throughout this work.
A binding is a single field in the configuration file; changing a provider requires editing that field and restarting the tool service, and never touches a Skill specification, the DEG format, or the Asset Registry.

\begin{table}[!t]
  \centering
  \caption{Provider bindings in the Provider and Configuration layer.
  Alternatives are selectable by editing the corresponding configuration fields.}
  \label{tab:provider_bindings}
  \fontsize{8}{9}\selectfont
  \setlength{\tabcolsep}{0.7mm}
  \renewcommand{\arraystretch}{1.2}
  \begin{tabular*}{\linewidth}{@{\extracolsep{\fill}}>{\raggedright\arraybackslash}p{0.21\linewidth} >{\raggedright\arraybackslash}p{0.31\linewidth} >{\raggedright\arraybackslash}p{0.26\linewidth} >{\raggedright\arraybackslash}p{0.19\linewidth}@{}}
    \toprule
    \textbf{Capability} & \textbf{Default provider (model/backend)} & \textbf{Alternative (model/backend)} & \textbf{API Key} \\
    \midrule
    \multicolumn{4}{c}{\cellcolor{catgray}\textit{\textbf{Understanding}}} \\
    \midrule
    Image & Host agent & --- & --- \\
    Video & Google (\texttt{gemini-2.5-flash}) & --- & \texttt{GEMINI\_API\_KEY} \\
    Speech & Google (\texttt{gemini-2.5-flash}) & OpenAI (\texttt{whisper-1}) & \texttt{GEMINI\_API\_KEY} / \texttt{OPENAI\_API\_KEY} \\
    Music and sound effects & Google (\texttt{gemini-2.5-flash}) & Alibaba (\texttt{Qwen2-Audio}) & \texttt{GEMINI\_API\_KEY} / \texttt{QWEN\_API\_KEY} \\
    Document & Mistral AI (\texttt{mistral-ocr-latest}) & Host agent & \texttt{MISTRAL\_API\_KEY} \\
    3D & Blender (\texttt{headless renderer}) & Local (\texttt{multi-view renderer}) & --- \\
    \midrule
    \multicolumn{4}{c}{\cellcolor{catgray}\textit{\textbf{Generation}}} \\
    \midrule
    Image & OpenAI (\texttt{gpt-image-2}) & fal.ai (\texttt{fal-ai/flux/schnell}) & \texttt{OPENAI\_API\_KEY} / \texttt{FAL\_API\_KEY} \\
    Video & Alibaba (\texttt{wan2.7}) & fal.ai (\texttt{Kling 1.6}) & \texttt{WAN\_API\_KEY} \\
    Music & ElevenLabs (\texttt{music\_v2}) & Replicate (\texttt{musicgen}) & \texttt{ELEVENLABS\_API\_KEY} / \texttt{REPLICATE\_API\_TOKEN} \\
    Sound effects & ElevenLabs (\texttt{text\_\allowbreak to\_\allowbreak sound\_\allowbreak v2}) & --- & \texttt{ELEVENLABS\_API\_KEY} \\
    Speech & OpenAI (\texttt{tts-1}) & Microsoft (\texttt{Edge TTS}) & \texttt{OPENAI\_API\_KEY} \\
    3D & Tripo AI (\texttt{v2.5}) & Meshy (\texttt{API v1/v2}) & \texttt{TRIPO\_API\_KEY} / \texttt{MESHY\_API\_KEY} \\
    PPT / Word / PDF / Excel & Local (\texttt{document renderers}) & --- & --- \\
    Code / Markdown & Host agent & --- & --- \\
    \midrule
    \multicolumn{4}{c}{\cellcolor{catgray}\textit{\textbf{Utility}}} \\
    \midrule
    Web search & Brave (\texttt{Search API}) & --- & \texttt{SEARCH\_API\_KEY} \\
    Web browsing & Local (\texttt{headless browser}) & --- & --- \\
    \bottomrule
  \end{tabular*}
\end{table}

\subsection{Default Parameters}
\label{app:mcp:defaults}

Each generation capability carries a set of defaults in the configuration layer.
A node in the DEG may override any of them through its \texttt{params} field, and the effective values are recorded in the resulting asset record, so that an asset can always be reproduced from what was registered.
Table~\ref{tab:defaults} lists the defaults used in this work.

\begin{table}[!t]
  \centering
  \caption{Default generation parameters.
  Node-level \texttt{params} override these values, and the effective values are persisted in the asset record as \texttt{params}.}
  \label{tab:defaults}
  \fontsize{8.5}{9.5}\selectfont
  \setlength{\tabcolsep}{3mm}
  \renewcommand{\arraystretch}{1.2}
  \begin{tabular}{@{}l l l@{}}
    \toprule
    \textbf{Capability} & \textbf{Parameter} & \textbf{Default} \\
    \midrule
    Image & \texttt{aspect\_ratio} / \texttt{quality} & \texttt{1:1} / \texttt{standard} \\
    Video & \texttt{duration\_seconds} / \texttt{aspect\_ratio} & \texttt{10} / \texttt{16:9} \\
    Sound effects & \texttt{duration\_seconds} & \texttt{15} \\
    Music & \texttt{duration\_seconds} / \texttt{instrumental} & \texttt{30} / \texttt{true} \\
    Speech & \texttt{language} / \texttt{speed} & \texttt{zh-CN} / \texttt{1.0} \\
    \bottomrule
  \end{tabular}
\end{table}

\subsection{Availability under Partial Configuration}
\label{app:mcp:availability}

Omni-IO Skills does not require a complete credential set.
Image understanding; PPT, Word, PDF, and Excel generation; code and Markdown generation; speech synthesis; and web browsing run natively in the host agent or locally, so they remain available without credentials.
On the first multimodal request, the host calls \texttt{check\_config}, which reports ready and missing capabilities, free alternatives when available, and capabilities that are always available.

\section{Skill Specification Format and Extensibility}
\label{app:skillspec}

\subsection{Shared Specification Schema}
\label{app:skillspec:schema}

Every Skill specification is a self-contained document loaded on demand by the Skill Entry layer.
Table~\ref{tab:skill_schema} lists the fields common to all three levels and the fields specific to each.
Levels differ in what they declare, not in how they are declared: an Atomic Skill declares the tool it invokes, an Expert Skill declares the workflow it expands into existing atomic tasks, and a Scenario Skill declares the rules by which it selects deliverables. None of them declares a service endpoint, a model name, or an API key.
\mbox{Figure~\ref{fig:atomic_skill} shows a specification in full.}
Specifications are written in the working language of their author; the specification in the figure is rendered in English for presentation.

\FloatBarrier

\begin{table}[t]
  \centering
  \caption{Fields of a Skill specification.
  The first block is shared by all three levels; the remaining blocks are level-specific.
  A specification is loaded only when the entry layer selects it.}
  \label{tab:skill_schema}
  \fontsize{8.5}{9.5}\selectfont
  \setlength{\tabcolsep}{2mm}
  \renewcommand{\arraystretch}{1.2}
  \begin{tabularx}{\linewidth}{@{}l >{\raggedright\arraybackslash}X@{}}
    \toprule
    \textbf{Field} & \textbf{Description} \\
    \midrule
    \multicolumn{2}{c}{\cellcolor{catgray}\textit{\textbf{Shared by all levels}}} \\
    \midrule
    Trigger condition & The request pattern or graph node type under which this Skill applies. \\
    Trigger boundary & The neighbouring cases that must \emph{not} route here, with the Skill that should handle them instead. \\
    Inputs & The fields consumed, their source, and whether each is required. \\
    Execution steps & The ordered procedure the host agent follows once the Skill is selected. \\
    Outputs & The asset object produced. \\
    Interactions & Which other Skills consume this Skill's output and what is injected downstream. \\
    Error handling & Per-error-class recovery, covering missing credentials, policy rejections, and timeouts. \\
    \midrule
    \multicolumn{2}{c}{\cellcolor{catgray}\textit{\textbf{Atomic Skills only}}} \\
    \midrule
    Tool binding & The configuration key and the MCP tool invoked, or the statement that execution is native. \\
    Parameter defaults & The configuration defaults applied when the node omits a parameter. \\
    \midrule
    \multicolumn{2}{c}{\cellcolor{catgray}\textit{\textbf{Expert Skills only}}} \\
    \midrule
    Expansion contract & The existing atomic task types the workflow expands into, and their dependency pattern. \\
    Assembly step & The node that combines the generated assets into the final artifact. \\
    Inspection criteria & What is checked on the assembled artifact and with which understanding capability. \\
    \midrule
    \multicolumn{2}{c}{\cellcolor{catgray}\textit{\textbf{Scenario Skills only}}} \\
    \midrule
    Deliverable rules & How the set of deliverables is inferred from the request, and the convention applied when it is underspecified. \\
    Level selection & For each deliverable, whether an Expert Skill or an Atomic Skill is invoked. \\
    Cross-deliverable reuse & Which deliverables share assets, and which may proceed in parallel. \\
    \bottomrule
  \end{tabularx}
\end{table}

\begin{figure}[!htbp]
\centering
\begin{minipage}{0.85\linewidth}
\begin{skillfigurebox}{image\_generation\_skill.md}
# Skill: Image Generation

## Trigger
A node of type "image" is present in the DEG.

## Tool binding
Configuration key : generate.image
MCP tool          : generate_image(prompt, aspect_ratio, quality)

## Inputs
prompt        <- node.prompt    complete, self-contained description of the image
aspect_ratio  <- node.params or configuration default
quality       <- node.params or configuration default

## Execution steps
1. Resolve parameters, filling omitted fields from the configuration defaults.
2. Invoke the bound provider through the MCP tool.
3. Write the returned image to the workspace output directory.
4. Register the asset and return the record.

## Outputs
An asset object of type "image" with an asset_id, a local path, a natural
language description, and the parameters that took effect.

## Interactions
downstream video : the local path is injected as the first video frame
downstream 3d    : the local path is injected as the reference image

## Error handling
missing credential : report the required key, mark the node failed
policy rejection   : report the rejected prompt, mark the node failed
network timeout    : retry once, then mark the node failed
\end{skillfigurebox}
\end{minipage}
\caption{A complete Atomic Skill specification of image generation.
The specification names a configuration key rather than a provider, so rebinding the provider in Table~\ref{tab:provider_bindings} leaves the specification unchanged.}
\label{fig:atomic_skill}
\end{figure}

\subsection{Routing Precedence and Invocation Direction}
\label{app:skillspec:routing}

Skill invocation is strictly downward. A Scenario Skill may invoke Expert or Atomic Skills, an Expert Skill may invoke Atomic Skills, and an Atomic Skill invokes no other Skill. Prohibiting upward and lateral calls keeps workflow expansion finite and the resulting DEG acyclic.

\subsection{Skills Extension}
\label{app:skillspec:extend}

Skill extension in Omni-IO Skills follows its hierarchical architecture. Atomic Skills add foundational executable capabilities, Expert Skills organize these capabilities into production workflows for concrete deliverables, and Scenario Skills compose lower-level Skills to support new application contexts. At all three levels, the extension is recorded in a specification that follows the format of Table~\ref{tab:skill_schema} and is registered in the Skill Entry index. The extension procedure for each level is described below.

\noindent\textbf{Atomic Skills extension.}
To add an Atomic Skill, the developer first encapsulates a capability that the host agent cannot execute natively as a new MCP tool and configures its provider binding.
The developer then defines the operation's trigger conditions, inputs, execution procedure, and outputs in the Skill specification and binds the specification to the new tool.

\noindent\textbf{Expert Skills extension.}
To add an Expert Skill, the developer specifies its target deliverable, decomposes the production workflow into existing Atomic Skills, and declares their dependencies.
The workflow also defines final artifact assembly, inspection, and localized revision.

\noindent\textbf{Scenario Skills extension.}
To add a Scenario Skill, the developer first defines its application context and trigger boundaries, then specifies the rules for selecting deliverables from a user request.
For each deliverable, the Scenario Skill selects an appropriate Expert or Atomic Skill and declares cross-deliverable dependencies, parallelism, and asset reuse.
A Scenario Skill coordinates existing lower-level Skills rather than introducing new executable operations.

\section{Expert Skill Production and Review Loop}
\label{app:expert}

\subsection{Input and Expansion Contracts}
\label{app:expert:contracts}

An Expert Skill receives the deliverable goal and its use context; the available inputs and reusable assets; content constraints whose accuracy must be preserved; the output specifications and production requirements; and the stylistic direction together with any exclusions.
It asks only for missing information that would materially affect the result and supplies defaults for adjustable details.

An Expert Skill distinguishes generative tasks from production operations that require precise control.
Content suited to generation is delegated to the corresponding Atomic Skills, whereas elements whose accuracy, structure, or timing must be guaranteed are handled through controllable content-processing or assembly steps.
This division depends on the deliverable rather than on a fixed production pipeline.

\begin{figure}[!b]
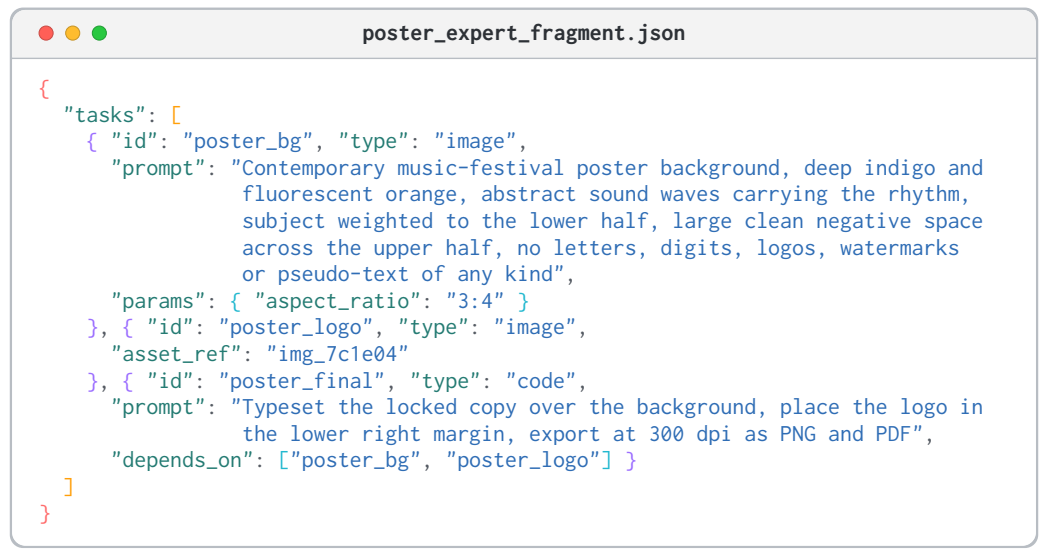

\centering
\begin{minipage}{0.85\linewidth}
\begin{jsonlistingbox}{poster\_expert\_fragment.json}
(*@\jsonoutertoken{\{}@*)
  (*@\jsonkeytoken{"tasks"}@*): (*@\jsonarraytoken{[}@*)
    (*@\jsonobjecttoken{\{}@*) (*@\jsonkeytoken{"id"}@*): "poster_bg", (*@\jsonkeytoken{"type"}@*): "image",
      (*@\jsonkeytoken{"prompt"}@*): "Contemporary music-festival poster background, deep indigo and
                 fluorescent orange, abstract sound waves carrying the rhythm,
                 subject weighted to the lower half, large clean negative space
                 across the upper half, no letters, digits, logos, watermarks
                 or pseudo-text of any kind",
      (*@\jsonkeytoken{"params"}@*): (*@\jsoninnertoken{\{}@*) (*@\jsonkeytoken{"aspect\_ratio"}@*): "3:4" (*@\jsoninnertoken{\}}@*)
    (*@\jsonobjecttoken{\}}@*), (*@\jsonobjecttoken{\{}@*) (*@\jsonkeytoken{"id"}@*): "poster_logo", (*@\jsonkeytoken{"type"}@*): "image",
      (*@\jsonkeytoken{"asset\_ref"}@*): "img_7c1e04"
    (*@\jsonobjecttoken{\}}@*), (*@\jsonobjecttoken{\{}@*) (*@\jsonkeytoken{"id"}@*): "poster_final", (*@\jsonkeytoken{"type"}@*): "code",
      (*@\jsonkeytoken{"prompt"}@*): "Typeset the locked copy over the background, place the logo in
                 the lower right margin, export at 300 dpi as PNG and PDF",
      (*@\jsonkeytoken{"depends\_on"}@*): (*@\jsoninnertoken{[}@*)"poster_bg", "poster_logo"(*@\jsoninnertoken{]}@*) (*@\jsonobjecttoken{\}}@*)
  (*@\jsonarraytoken{]}@*)
(*@\jsonoutertoken{\}}@*)
\end{jsonlistingbox}
\end{minipage}
\caption{An Expert Skill expansion fragment for a poster.
Only existing task types appear: the visual asset is an \texttt{image} node, a registered logo enters as an asset-source node, and assembly is a \texttt{code} node that depends on both.
Task identifiers name the deliverable so that fragments from several Expert Skills can be merged without collision.}
\label{fig:expert_fragment}
\end{figure}

An Expert Skill is not itself an execution unit.
It decomposes the complete production workflow into existing executable tasks, declares their dependencies, and emits a flat task fragment.
The fragment is merged with the rest of the graph and validated and scheduled by the common execution procedure.
The two implemented Expert Skills instantiate this contract through deliverable-specific production stages and modality-appropriate inspection, as summarized in Table~\ref{tab:expert_workflows}.
Figure~\ref{fig:expert_fragment} illustrates this expansion for poster production.

\begin{table}[!t]
  \centering
  \caption{Implemented Expert Skills.
  Each expands into existing atomic task types, assembles the final artifact through a \texttt{code} node, and inspects the result with the understanding capability appropriate to the artifact's modality.}
  \label{tab:expert_workflows}
  \fontsize{8.5}{9.5}\selectfont
  \setlength{\tabcolsep}{2mm}
  \renewcommand{\arraystretch}{1.2}
  \begin{tabularx}{\linewidth}{@{}l >{\raggedright\arraybackslash}X >{\raggedright\arraybackslash}X@{}}
    \toprule
    \textbf{Expert Skill} & \textbf{Workflow stages} & \textbf{Inspection} \\
    \midrule
    Poster Design & Lock the brief and exact copy $\rightarrow$ fix the visual direction $\rightarrow$ generate only the necessary visual assets $\rightarrow$ typeset copy and assets deterministically $\rightarrow$ inspect. & Native vision on the assembled image, checking copy fidelity, hierarchy, and legibility. \\
    Complex Video Production & Script and storyboard $\rightarrow$ generate shots, narration, and music $\rightarrow$ assemble with subtitles and branding $\rightarrow$ inspect. & Video understanding on the assembled cut, with audio understanding where narration or music must be verified. \\
    \bottomrule
  \end{tabularx}
\end{table}

\subsection{Assembly}
\label{app:expert:assembly}

Assembly uses a controllable programmatic workflow to combine generated content and existing assets into the final artifact.
A poster is typeset programmatically over its generated background so that copy is rendered with exact glyphs rather than sampled from a generative model; a video is cut, mixed, and subtitled by a local media pipeline.

\subsection{Inspection and Localized Rework}
\label{app:expert:review}

An Expert Skill does not deliver on the completion of its atomic nodes.
The assembled artifact is read back with the understanding capability of its own modality and checked against the locked brief.
When a defect is found, it is localized to the smallest responsible part of the workflow, and only that part is reworked.
Table~\ref{tab:rework} summarizes the corresponding actions.

\begin{table}[!t]
  \centering
  \caption{Defect attribution and the corresponding minimal rework.
  Validated assets are reused across rework rounds, so a localized defect never triggers regeneration of the whole artifact.}
  \label{tab:rework}
  \fontsize{8.5}{9.5}\selectfont
  \setlength{\tabcolsep}{2mm}
  \renewcommand{\arraystretch}{1.2}
  \begin{tabularx}{\linewidth}{@{}l >{\raggedright\arraybackslash}X >{\raggedright\arraybackslash}X@{}}
    \toprule
    \textbf{Responsible layer} & \textbf{Symptom} & \textbf{Minimal rework} \\
    \midrule
    Atomic asset & A generated visual or shot is off-brief, malformed, or carries spurious text. & Re-run that atomic node only, then re-assemble. \\
    Written content & Copy, narration, or lyrics misstate a locked fact. & Re-run the affected content node, then re-assemble. \\
    Assembly & Layout, cut, subtitle timing, transition, or mix is wrong while the assets are correct. & Modify the \texttt{code} node only; no generation call is issued. \\
    Planning & The deliverable itself was misconceived or the brief was incomplete. & Return to the user with the specific decision required. \\
    \bottomrule
  \end{tabularx}
\end{table}

After rework, assembly must use the newly generated assets rather than continue referencing rejected versions.
Each new artifact is inspected again.
Autonomous rework is limited to two rounds by default; if the issue remains unresolved, the Expert Skill reports the outstanding problem, delivers the best available version, and asks the user for the smallest decision needed to resolve it.

\section{Scenario Skill Deliverable Planning and Coordination}
\label{app:scenario}

\subsection{Input and Planning Contracts}
\label{app:scenario:scope}

A Scenario Skill receives the user's objective, source materials, reusable assets, explicitly requested deliverables, and constraints shared across outputs.
It first records the requirements shared by all deliverables in a common scenario brief, then prepares a separate production brief for the specific requirements of each deliverable.
Explicit user scope is preserved; when the output form is unspecified, the Skill selects the smallest conventional set that satisfies the objective.
It asks for clarification only when ambiguity would materially affect scope, cost, or intended use.
\label{app:scenario:selection}
Table~\ref{tab:scenario_deliverables} lists the candidate outputs of the six implemented Scenario Skills.

\begin{table}[H]
  \centering
  \caption{Deliverable-selection domains of the implemented Scenario Skills.}
  \label{tab:scenario_deliverables}
  \fontsize{8.5}{9.5}\selectfont
  \setlength{\tabcolsep}{2mm}
  \renewcommand{\arraystretch}{1.2}
  \begin{tabularx}{\linewidth}{@{}l >{\raggedright\arraybackslash}X >{\raggedright\arraybackslash}X@{}}
    \toprule
    \textbf{Scenario Skill} & \textbf{Application boundary} & \textbf{Candidate deliverables} \\
    \midrule
    Social-Media Post & Platform-facing publication or promotion around supplied or generated source material. & Post copy, supporting image or poster, short video, and optional audio. \\
    Office Documents & Transformation of source material into structured workplace artifacts. & Presentation, report, PDF, minutes, and spreadsheet. \\
    Job Application & A coordinated application package for a target role. & Resume, cover letter, self-introduction presentation, speech, and video. \\
    Education Sharing & Teaching or explanation for a specified audience and learning objective. & Slides, handout, illustration, narration, and explainer video. \\
    Event Material & Promotion, participation, or retrospective material for one event. & Invitation, poster, atmosphere music, teaser video, and retrospective video. \\
    Game Asset & A coordinated presentation of one game concept or production asset. & Concept art, 3D model, showcase video, and optional web presentation. \\
    \bottomrule
  \end{tabularx}
\end{table}

\label{app:scenario:expansion}

For each deliverable, the Scenario Skill chooses the lowest Skill level capable of completing the required procedure.
Tasks that require coordinated asset generation, deterministic assembly, artifact-level inspection, and localized rework are delegated to an Expert Skill; tasks that can be completed as one independent operation are delegated directly to an Atomic Skill.
The expanded task fragments are merged into the DEG, and shared constraints are written into every relevant brief.

\label{app:scenario:example}

\begin{figure}[!t]
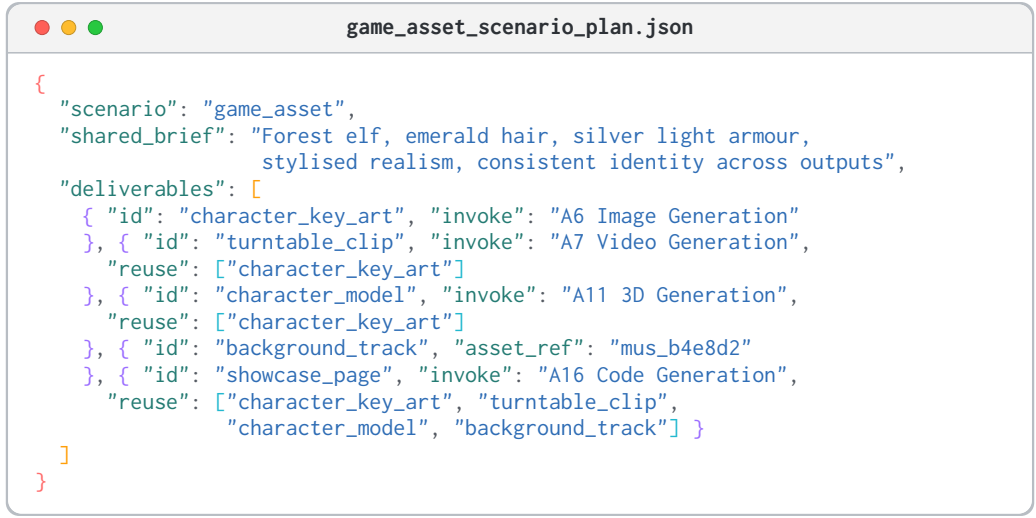

\centering
\begin{minipage}{0.85\linewidth}
\begin{jsonlistingbox}{game\_asset\_scenario\_plan.json}
(*@\jsonoutertoken{\{}@*)
  (*@\jsonkeytoken{"scenario"}@*): "game_asset",
  (*@\jsonkeytoken{"shared\_brief"}@*): "Forest elf, emerald hair, silver light armour,
                   stylised realism, consistent identity across outputs",
  (*@\jsonkeytoken{"deliverables"}@*): (*@\jsonarraytoken{[}@*)
    (*@\jsonobjecttoken{\{}@*) (*@\jsonkeytoken{"id"}@*): "character_key_art", (*@\jsonkeytoken{"invoke"}@*): "A6 Image Generation"
    (*@\jsonobjecttoken{\}}@*), (*@\jsonobjecttoken{\{}@*) (*@\jsonkeytoken{"id"}@*): "turntable_clip", (*@\jsonkeytoken{"invoke"}@*): "A7 Video Generation",
      (*@\jsonkeytoken{"reuse"}@*): (*@\jsoninnertoken{[}@*)"character_key_art"(*@\jsoninnertoken{]}@*)
    (*@\jsonobjecttoken{\}}@*), (*@\jsonobjecttoken{\{}@*) (*@\jsonkeytoken{"id"}@*): "character_model", (*@\jsonkeytoken{"invoke"}@*): "A11 3D Generation",
      (*@\jsonkeytoken{"reuse"}@*): (*@\jsoninnertoken{[}@*)"character_key_art"(*@\jsoninnertoken{]}@*)
    (*@\jsonobjecttoken{\}}@*), (*@\jsonobjecttoken{\{}@*) (*@\jsonkeytoken{"id"}@*): "background_track", (*@\jsonkeytoken{"asset\_ref"}@*): "mus_b4e8d2"
    (*@\jsonobjecttoken{\}}@*), (*@\jsonobjecttoken{\{}@*) (*@\jsonkeytoken{"id"}@*): "showcase_page", (*@\jsonkeytoken{"invoke"}@*): "A16 Code Generation",
      (*@\jsonkeytoken{"reuse"}@*): (*@\jsoninnertoken{[}@*)"character_key_art", "turntable_clip",
                "character_model", "background_track"(*@\jsoninnertoken{]}@*) (*@\jsonobjecttoken{\}}@*)
  (*@\jsonarraytoken{]}@*)
(*@\jsonoutertoken{\}}@*)
\end{jsonlistingbox}
\end{minipage}
\caption{A Scenario Skill deliverable plan for a forest-elf game-asset package.
The plan records lower-level Skill invocations and asset-reuse relationships before being expanded into a DEG.}
\label{fig:scenario_plan}
\end{figure}

Consider a forest-elf game-asset request containing a character illustration, a short turntable clip, a 3D model, and a single-page showcase, with an existing background track reused.
Figure~\ref{fig:scenario_plan} shows the resulting deliverable plan: the illustration serves as the shared identity anchor; the video and 3D model run in parallel after it completes; the background track is reused directly; and the showcase page aggregates all assets at the end.

\subsection{Cross-Deliverable Coordination and Reuse}
\label{app:scenario:coordination}

The common scenario brief is the consistency anchor for the package.
Locked facts, names, dates, claims, brand requirements, and exclusions are propagated unchanged to every affected deliverable.
Stylistic direction is specialized only where the output medium requires it, without changing the common identity of the package.
For example, a social-media caption and a promotional video may use different lengths and pacing while preserving the same product facts and campaign direction.

Source analysis and generated assets are represented once whenever several deliverables consume the same result.
If a downstream output must read or transform an upstream artifact, the dependency is declared explicitly.
If two outputs share only the common scenario brief and consume no common artifact, their prompts are made self-contained and the branches may proceed in parallel.
This distinction preserves both asset identity and the parallelism available to the Wave scheduler.

\subsection{Completion and Revision}
\label{app:scenario:completion}

After execution, the Scenario Skill performs a package-level completion and check.
It verifies that every selected deliverable has a terminal state, that completed outputs preserve the locked facts and declared relationships, and that intended reuse points reference the selected asset versions. 
If a shared dependency fails, dependent deliverables become unavailable while independent branches may still complete.
The Scenario Skill reports the package as complete, partial, or failed and lists any unavailable deliverables.
User revisions are applied according to their scope.
Local changes affect only the target deliverable and its descendants, while changes to locked facts, shared assets, or global direction update all affected briefs and dependencies.
Completed assets remain available, and revised branches register new versions.

\section{Declare Execution Graph Specification}
\label{app:deg}

\subsection{Reference Orchestration Procedure}
\label{app:orchestration_algorithm}

Algorithm~\ref{alg:orchestration} records the reference orchestration procedure underlying Figure~\ref{fig:runtime_workflow}.
The procedure makes explicit how Omni-IO Skills expansion produces terminal tasks, how a DEG is validated and scheduled in parallel Waves, and how successful outputs and failure states are propagated through the Asset Registry.

\begin{algorithm}[!t]
\small
\caption{Dependency-aware orchestration in Omni-IO Skills.}
\label{alg:orchestration}
\begin{algorithmic}[1]
\Require user request $q$, Skill index $\mathcal{S}$, Asset Registry $\mathcal{R}$
\Ensure outputs and final execution states of all nodes
\State $\mathcal{T} \gets \textsc{SkillEntry}(q, \mathcal{S})$ \Comment{selected Skills expanded into terminal tasks}
\State $G = (V, E)$ with $V \gets \mathcal{T}$ and $E$ induced by \texttt{depends\_on}
\If{$\textsc{Validate}(G, \mathcal{R})$ fails}
  \State \Return planning error \Comment{unresolved reference or dependency cycle}
\EndIf
\State $\mathrm{status}[v] \gets \textsc{Pending}$ for all $v \in V$
\ForAll{$v \in V$ with \texttt{asset\_ref} present}
  \State $o[v] \gets \mathcal{R}.\textsc{Get}(v.\texttt{asset\_ref})$; $\mathrm{status}[v] \gets \textsc{Completed}$
\EndFor
\While{some $v \in V$ has $\mathrm{status}[v]=\textsc{Pending}$}
  \State $\mathcal{W} \gets \{\, v \in V : \mathrm{status}[v]=\textsc{Pending} \text{ and all predecessors completed} \,\}$
  \ForAll{$v \in \mathcal{W}$ \textbf{in parallel}}
    \State $\pi \gets \textsc{ResolveProvider}(v.\texttt{type})$
    \State $\theta \gets \pi.\mathrm{defaults} \oplus v.\texttt{params} \oplus \textsc{Inject}\big(o[\mathrm{pred}(v)]\big)$
    \State $o[v] \gets \textsc{Execute}(\pi, v.\texttt{prompt}, \theta)$
    \If{execution succeeded}
      \State $\mathcal{R}.\textsc{Register}(o[v])$; $\mathrm{status}[v] \gets \textsc{Completed}$
    \Else
      \State $\mathrm{status}[v] \gets \textsc{Failed}$
      \State mark all pending descendants cancelled \Comment{independent nodes continue}
    \EndIf
  \EndFor
\EndWhile
\State \Return $(o, \mathrm{status})$
\end{algorithmic}
\end{algorithm}

The procedure maintains separate control and asset states.
Node status determines which tasks become schedulable or cancelled, while the registry records only successfully materialized artifacts.
This separation allows independent graph branches to continue after a local failure and makes completed assets available to later nodes and future turns.

\subsection{Node Schema}
\label{app:deg:schema}

Table~\ref{tab:node_schema} lists the fields of a DEG node.
A node is either a generation node carrying a \texttt{prompt} or an asset-source node carrying an \texttt{asset\_ref}.
The two fields are mutually exclusive.
Prompts are self-contained and describe everything a node needs without referring to another node's prompt.
Semantic consistency across concurrent nodes is established by writing their prompts within the same planning context.

\begin{table}[!t]
  \centering
  \captionsetup{skip=4pt}
  \caption{Fields of a DEG node.
  An asterisk marks the two mutually exclusive fields.
  Exactly one of \texttt{prompt} and \texttt{asset\_ref} is present.
  Upstream file paths are supplied by the scheduler at invocation time, so a prompt never contains a path or an asset identifier.}
  \label{tab:node_schema}
  \fontsize{8.5}{9.5}\selectfont
  \setlength{\tabcolsep}{2mm}
  \renewcommand{\arraystretch}{1.1}
  \begin{tabularx}{\linewidth}{@{}l l c >{\raggedright\arraybackslash}X@{}}
    \toprule
    \textbf{Field} & \textbf{Type} & \textbf{Req.} & \textbf{Meaning} \\
    \midrule
    \texttt{id} & string & \cmark & Unique identifier within the graph; the target of \texttt{depends\_on}. \\
    \texttt{type} & string & \cmark & Task type, drawn from Table~\ref{tab:task_types}. \\
    \texttt{prompt} & string & $\ast$ & Complete, self-contained description of what this node must produce. \\
    \texttt{params} & object & \xmark & Parameter overrides applied on top of the configuration defaults. \\
    \texttt{depends\_on} & string \textbar{} array & \xmark & Upstream node identifiers; an array is admissible only for \texttt{code} and \texttt{markdown}. \\
    \texttt{asset\_ref} & string & $\ast$ & Identifier of a previously registered asset; makes this an asset-source node. \\
    \bottomrule
  \end{tabularx}
\end{table}

\subsection{Task Types and Execution Sites}
\label{app:deg:types}

Table~\ref{tab:task_types} lists the admissible node types.
Most types correspond to one MCP tool call with a fixed parameter signature, and therefore cannot accept a variable number of upstream assets; \texttt{code} and \texttt{markdown} are written natively by the host agent, which resolves each upstream identifier through \texttt{get\_asset} before writing, and are consequently the only types for which a dependency array is meaningful.

\begin{table}[!t]
  \centering
  \captionsetup{skip=4pt}
  \caption{Admissible node types and where they execute.
  Only the two natively executed types can consume multiple upstream assets, and only they require explicit registration.}
  \label{tab:task_types}
  \fontsize{8.5}{9.5}\selectfont
  \setlength{\tabcolsep}{2.5mm}
  \renewcommand{\arraystretch}{1.1}
  \begin{tabular}{@{}l l c c@{}}
    \toprule
    \textbf{Type} & \textbf{Execution site} & \textbf{Multi-dependency} & \textbf{Auto-registered} \\
    \midrule
    \texttt{image}, \texttt{video}, \texttt{3d} & MCP tool & \xmark & \cmark \\
    \texttt{music}, \texttt{sfx}, \texttt{speech} & MCP tool & \xmark & \cmark \\
    \texttt{ppt}, \texttt{word}, \texttt{pdf}, \texttt{excel} & MCP tool (local renderer) & \xmark & \cmark \\
    \texttt{code}, \texttt{markdown} & Host agent native file writing & \cmark & \xmark \\
    existing asset (\texttt{asset\_ref} present) & None; resolved from the registry & \xmark & --- \\
    \bottomrule
  \end{tabular}
\end{table}

\subsection{Validation Rules}
\label{app:deg:validation}

Before any tool is invoked, the graph is checked against the rules in Table~\ref{tab:validation}.
Each rule imposes a structural requirement on the declaration that is decidable without executing anything.
A graph that violates any rule is a planning error and is reconstructed rather than submitted for execution.
When it is unclear whether a dependency is required, the system declares it.

\begin{table}[!t]
  \centering
  \captionsetup{skip=4pt}
  \caption{Validation rules applied to the DEG before execution.
  Every rule is decidable from the declaration alone.
  Failure returns the graph to planning rather than to execution.}
  \label{tab:validation}
  \fontsize{8.5}{9.5}\selectfont
  \setlength{\tabcolsep}{2mm}
  \renewcommand{\arraystretch}{1.05}
  \begin{tabularx}{\linewidth}{@{}l >{\raggedright\arraybackslash}X >{\raggedright\arraybackslash}X@{}}
    \toprule
    \textbf{Rule} & \textbf{What it demands} & \textbf{What it prevents} \\
    \midrule
    Identifier uniqueness & Every \texttt{id} occurs once in the graph. & A dependency resolving to the wrong node. \\
    Type admissibility & Every \texttt{type} appears in Table~\ref{tab:task_types}. & An Expert Skill inventing a node type that no execution site can run. \\
    Node completeness & Exactly one of \texttt{prompt} and \texttt{asset\_ref} is present, and a \texttt{prompt} is self-contained. & A generation call issued with an empty or context-dependent description. \\
    Dependency resolution & Every identifier in \texttt{depends\_on} names a node in the same graph. & A node waiting forever on a task that was never declared. \\
    Asset resolution & Every \texttt{asset\_ref} names a record in the registry. & An asset-source node resolving to a path that does not exist. \\
    Acyclicity & The graph induced by \texttt{depends\_on} contains no cycle. & A Wave partition that never terminates. \\
    Dependency arity & An array \texttt{depends\_on} occurs only on \texttt{code} and \texttt{markdown} nodes. & A fixed-signature tool call being handed a variable number of upstream assets. \\
    Conservative declaration & A dependency is declared whenever a node consumes an upstream output, references its content, or must remain semantically consistent with it. & A downstream node starting before the asset it depends on exists. \\
    \bottomrule
\end{tabularx}
\end{table}

\subsection{Failure and Interruption Semantics}
\label{app:deg:failure}

Node failure and user interruption both cancel affected downstream work without rebuilding the graph.
A failed node is marked failed, its pending descendants are cancelled, and independent branches continue.
Failed and cancelled nodes are reported but not registered.

User instructions modify the graph in flight.
For each pending or in-flight node, the host agent determines whether its output is affected.
Affected in-flight nodes are cancelled and re-issued, affected pending nodes are edited before dispatch, and unaffected nodes continue.
Completed assets remain registered.
Table~\ref{tab:interruption} summarizes the handling of five interruption classes.

\begin{table}[H]
  \centering
  \captionsetup{skip=4pt}
  \caption{Interruption classes and their handling.
  Completed assets remain registered; only pending and in-flight nodes are affected.}
  \label{tab:interruption}
  \fontsize{8.5}{9.5}\selectfont
  \setlength{\tabcolsep}{2mm}
  \renewcommand{\arraystretch}{1.05}
  \begin{tabularx}{\linewidth}{@{}l >{\hsize=.75\hsize\linewidth=\hsize\raggedright\arraybackslash}X >{\hsize=1.25\hsize\linewidth=\hsize\raggedright\arraybackslash}X@{}}
    \toprule
    \textbf{Class} & \textbf{Recognized by} & \textbf{Handling} \\
    \midrule
    Local revision & One deliverable changes. & Update the target; pending descendants wait for its new output, while completed descendants remain linked to the prior version. \\
    Global revision & A shared scene, style, or tone changes. & Cancel and re-issue affected in-flight nodes; regenerate completed assets only on request. \\
    Extension & A deliverable is added. & Insert new nodes; dispatch independent nodes immediately and schedule dependent nodes in later Waves. \\
    Cancellation & A deliverable is withdrawn. & Cancel the node and its pending descendants; register no output for them. \\
    Topic switch & The instruction is unrelated to the graph. & Confirm, cancel remaining work, and plan a new graph; keep completed assets. \\
    \bottomrule
  \end{tabularx}
\end{table}

\begin{table}[!t]
  \centering
  \captionsetup{skip=4pt}
  \caption{Wave partition of Figure~\ref{fig:deg_example}.
  Nodes in the same Wave are mutually independent and are dispatched in one concurrent round.
  The asset-source node is complete on entry and issues no call.}
  \label{tab:waves}
  \fontsize{8.5}{9.5}\selectfont
  \setlength{\tabcolsep}{3mm}
  \renewcommand{\arraystretch}{1.05}
  \begin{tabular}{@{}l l l@{}}
    \toprule
    \textbf{Wave} & \textbf{Nodes} & \textbf{Calls issued} \\
    \midrule
    0 & \texttt{img1}, \texttt{mus1} & one image generation; \texttt{mus1} resolves from the registry \\
    1 & \texttt{vid1}, \texttt{3d1} & one video and one 3D generation, concurrently \\
    2 & \texttt{code1} & none; written natively after resolving four asset paths \\
    \bottomrule
  \end{tabular}
\end{table}

\begin{figure}[!t]
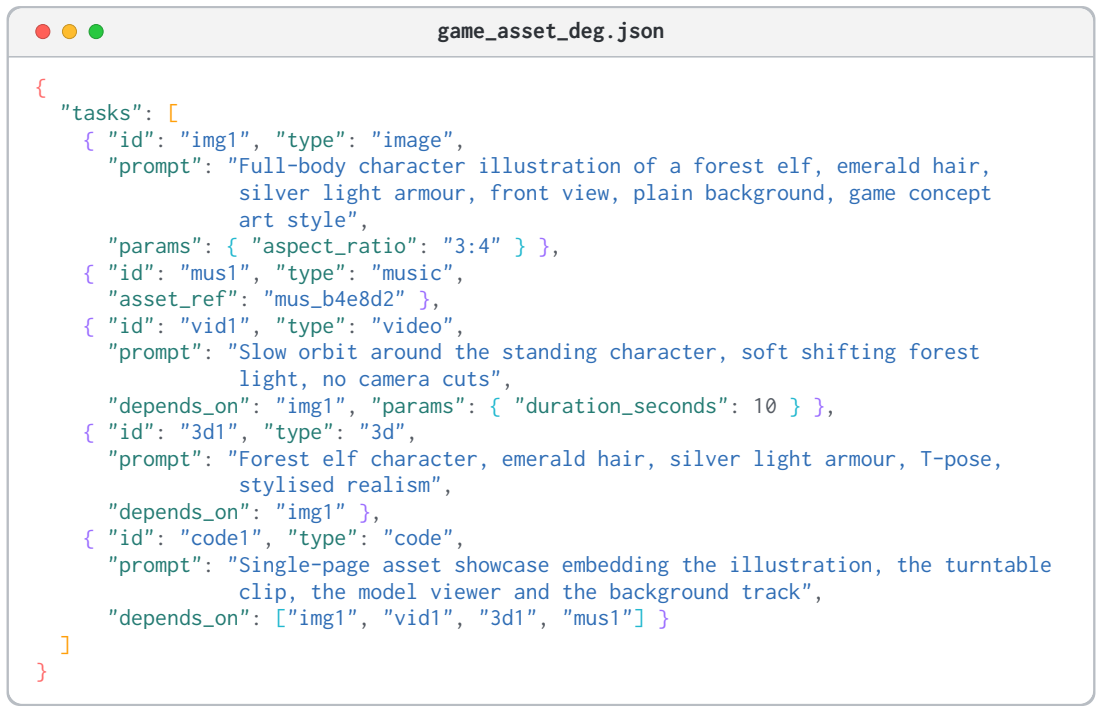

\centering
\begin{minipage}{0.90\linewidth}
\begin{jsonlistingbox}{game\_asset\_deg.json}
(*@\jsonoutertoken{\{}@*)
  (*@\jsonkeytoken{"tasks"}@*): (*@\jsonarraytoken{[}@*)
    (*@\jsonobjecttoken{\{}@*) (*@\jsonkeytoken{"id"}@*): "img1", (*@\jsonkeytoken{"type"}@*): "image",
      (*@\jsonkeytoken{"prompt"}@*): "Full-body character illustration of a forest elf, emerald hair,
                 silver light armour, front view, plain background, game concept
                 art style",
      (*@\jsonkeytoken{"params"}@*): (*@\jsoninnertoken{\{}@*) (*@\jsonkeytoken{"aspect\_ratio"}@*): "3:4" (*@\jsoninnertoken{\}}@*) (*@\jsonobjecttoken{\}}@*),
    (*@\jsonobjecttoken{\{}@*) (*@\jsonkeytoken{"id"}@*): "mus1", (*@\jsonkeytoken{"type"}@*): "music",
      (*@\jsonkeytoken{"asset\_ref"}@*): "mus_b4e8d2" (*@\jsonobjecttoken{\}}@*),
    (*@\jsonobjecttoken{\{}@*) (*@\jsonkeytoken{"id"}@*): "vid1", (*@\jsonkeytoken{"type"}@*): "video",
      (*@\jsonkeytoken{"prompt"}@*): "Slow orbit around the standing character, soft shifting forest
                 light, no camera cuts",
      (*@\jsonkeytoken{"depends\_on"}@*): "img1", (*@\jsonkeytoken{"params"}@*): (*@\jsoninnertoken{\{}@*) (*@\jsonkeytoken{"duration\_seconds"}@*): 10 (*@\jsoninnertoken{\}}@*) (*@\jsonobjecttoken{\}}@*),
    (*@\jsonobjecttoken{\{}@*) (*@\jsonkeytoken{"id"}@*): "3d1", (*@\jsonkeytoken{"type"}@*): "3d",
      (*@\jsonkeytoken{"prompt"}@*): "Forest elf character, emerald hair, silver light armour, T-pose,
                 stylised realism",
      (*@\jsonkeytoken{"depends\_on"}@*): "img1" (*@\jsonobjecttoken{\}}@*),
    (*@\jsonobjecttoken{\{}@*) (*@\jsonkeytoken{"id"}@*): "code1", (*@\jsonkeytoken{"type"}@*): "code",
      (*@\jsonkeytoken{"prompt"}@*): "Single-page asset showcase embedding the illustration, the turntable
                 clip, the model viewer and the background track",
      (*@\jsonkeytoken{"depends\_on"}@*): (*@\jsoninnertoken{[}@*)"img1", "vid1", "3d1", "mus1"(*@\jsoninnertoken{]}@*) (*@\jsonobjecttoken{\}}@*)
  (*@\jsonarraytoken{]}@*)
(*@\jsonoutertoken{\}}@*)
\end{jsonlistingbox}
\end{minipage}
\caption{A DEG combining cross-turn reuse, parallel generation, and aggregation.
The video and 3D nodes both consume the character illustration, and the page node aggregates all four assets.}
\label{fig:deg_example}
\end{figure}

\subsection{Worked Example}
\label{app:deg:example}

Figure~\ref{fig:deg_example} declares a graph for a request that combines a reused asset, two independent generations, and an aggregation, and Table~\ref{tab:waves} gives the resulting Wave partition.
Two properties are visible.
The asset-source node occupies Wave~0 as an already completed node and issues no generation call, so reusing a previous turn's output costs nothing.
The two nodes of Wave~1 are mutually independent and are dispatched in a single concurrent round, and their failure modes are independent: were the 3D node to fail, the video node would still complete, and only the aggregation node would be cancelled.

\section{Asset Registry Schema and Cross-Turn Reuse}
\label{app:registry}

\subsection{Record Schema}
\label{app:registry:schema}

Table~\ref{tab:registry_schema} lists the fields of a registry record.
The \texttt{description} field is written at generation time and supports semantic retrieval in subsequent turns.
The \texttt{turn\_id}, set once per user message, disambiguates between assets of the same type produced in different turns.

\begin{table}[!t]
  \centering
  \captionsetup{skip=4pt}
  \caption{Fields of an Asset Registry record.
  Records are keyed by \texttt{asset\_id} and persisted as a single JSON file in the workspace.}
  \label{tab:registry_schema}
  \fontsize{8.5}{9.5}\selectfont
  \setlength{\tabcolsep}{2mm}
  \renewcommand{\arraystretch}{1.05}
  \begin{tabularx}{\linewidth}{@{}l >{\raggedright\arraybackslash}X@{}}
    \toprule
    \textbf{Field} & \textbf{Meaning} \\
    \midrule
    \texttt{asset\_id} & Globally unique identifier, formed from a type prefix and a random suffix. \\
    \texttt{type} & \texttt{image}, \texttt{video}, \texttt{audio}, \texttt{3d}, \texttt{document}, or \texttt{code}. \\
    \texttt{subtype} & Finer classification: \texttt{music}, \texttt{sfx}, \texttt{speech}; \texttt{ppt}, \texttt{word}, \texttt{pdf}, \texttt{excel}, \texttt{markdown}; \texttt{poster}, \texttt{produced\_video} for assembled artifacts. \\
    \texttt{path} & Absolute local path of the file in the workspace output directory. \\
    \texttt{description} & Natural language description written at generation time, used for semantic retrieval in later turns. \\
    \texttt{params} & The parameters that actually took effect, after defaults and node-level overrides were merged. \\
    \texttt{turn\_id} & Identifier of the interaction turn, set once per user message. \\
    \texttt{source\_asset\_id} & Optional lineage pointer to the record this asset was derived from. \\
    \bottomrule
  \end{tabularx}
\end{table}

\subsection{Registration Paths}
\label{app:registry:paths}

Assets reach the registry along two paths.
Outputs of MCP generation tools are registered by the tool service before the call returns, so the host agent performs no bookkeeping.
Files written natively by the host agent, namely \texttt{code} and \texttt{markdown} nodes and the artifacts assembled by Expert Skills, are registered by an explicit call after the file is written.

\subsection{Concurrency and Versioning}
\label{app:registry:concurrency}

The registry is a JSON file in the workspace, shared across sessions.
Writes are atomic, performed to a temporary file and renamed into place, and serialized with a file lock on platforms that provide one, so that concurrent sessions registering into the same workspace do not overwrite one another.

Records are appended, never updated in place.
Regenerating a deliverable produces a new record with a new \texttt{asset\_id} that may point back at the record it was derived from, so successive versions of the same artifact remain distinguishable and individually retrievable.

\subsection{Cross-Turn Reuse}
\label{app:registry:reuse}

\begin{figure}[!t]
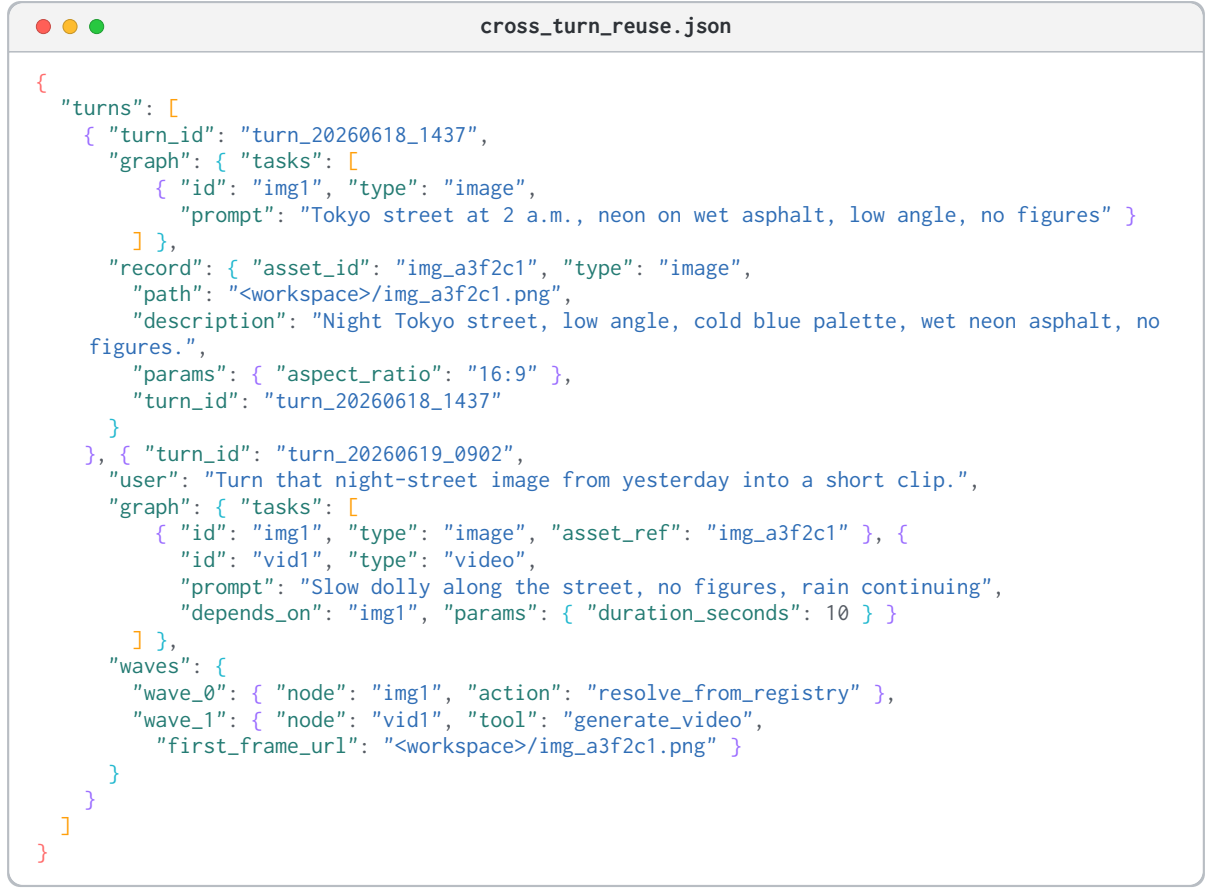

\centering
\begin{minipage}{0.99\linewidth}
\begin{jsonlistingbox}{cross\_turn\_reuse.json}
(*@\jsonoutertoken{\{}@*)
  (*@\jsonkeytoken{"turns"}@*): (*@\jsonarraytoken{[}@*)
    (*@\jsonobjecttoken{\{}@*) (*@\jsonkeytoken{"turn\_id"}@*): "turn_20260618_1437",
      (*@\jsonkeytoken{"graph"}@*): (*@\jsoninnertoken{\{}@*) (*@\jsonkeytoken{"tasks"}@*): (*@\jsonarraytoken{[}@*)
          (*@\jsonobjecttoken{\{}@*) (*@\jsonkeytoken{"id"}@*): "img1", (*@\jsonkeytoken{"type"}@*): "image",
            (*@\jsonkeytoken{"prompt"}@*): "Tokyo street at 2 a.m., neon on wet asphalt, low angle, no figures" (*@\jsonobjecttoken{\}}@*)
        (*@\jsonarraytoken{]}@*) (*@\jsoninnertoken{\}}@*),
      (*@\jsonkeytoken{"record"}@*): (*@\jsoninnertoken{\{}@*) (*@\jsonkeytoken{"asset\_id"}@*): "img_a3f2c1", (*@\jsonkeytoken{"type"}@*): "image",
        (*@\jsonkeytoken{"path"}@*): "<workspace>/img_a3f2c1.png",
        (*@\jsonkeytoken{"description"}@*): "Night Tokyo street, low angle, cold blue palette, wet neon asphalt, no figures.",
        (*@\jsonkeytoken{"params"}@*): (*@\jsonobjecttoken{\{}@*) (*@\jsonkeytoken{"aspect\_ratio"}@*): "16:9" (*@\jsonobjecttoken{\}}@*),
        (*@\jsonkeytoken{"turn\_id"}@*): "turn_20260618_1437"
      (*@\jsoninnertoken{\}}@*)
    (*@\jsonobjecttoken{\}}@*), (*@\jsonobjecttoken{\{}@*) (*@\jsonkeytoken{"turn\_id"}@*): "turn_20260619_0902",
      (*@\jsonkeytoken{"user"}@*): "Turn that night-street image from yesterday into a short clip.",
      (*@\jsonkeytoken{"graph"}@*): (*@\jsoninnertoken{\{}@*) (*@\jsonkeytoken{"tasks"}@*): (*@\jsonarraytoken{[}@*)
          (*@\jsonobjecttoken{\{}@*) (*@\jsonkeytoken{"id"}@*): "img1", (*@\jsonkeytoken{"type"}@*): "image", (*@\jsonkeytoken{"asset\_ref"}@*): "img_a3f2c1" (*@\jsonobjecttoken{\}}@*), (*@\jsonobjecttoken{\{}@*)
            (*@\jsonkeytoken{"id"}@*): "vid1", (*@\jsonkeytoken{"type"}@*): "video",
            (*@\jsonkeytoken{"prompt"}@*): "Slow dolly along the street, no figures, rain continuing",
            (*@\jsonkeytoken{"depends\_on"}@*): "img1", (*@\jsonkeytoken{"params"}@*): (*@\jsoninnertoken{\{}@*) (*@\jsonkeytoken{"duration\_seconds"}@*): 10 (*@\jsoninnertoken{\}}@*) (*@\jsonobjecttoken{\}}@*)
        (*@\jsonarraytoken{]}@*) (*@\jsoninnertoken{\}}@*),
      (*@\jsonkeytoken{"waves"}@*): (*@\jsoninnertoken{\{}@*)
        (*@\jsonkeytoken{"wave\_0"}@*): (*@\jsonobjecttoken{\{}@*) (*@\jsonkeytoken{"node"}@*): "img1", (*@\jsonkeytoken{"action"}@*): "resolve_from_registry" (*@\jsonobjecttoken{\}}@*),
        (*@\jsonkeytoken{"wave\_1"}@*): (*@\jsonobjecttoken{\{}@*) (*@\jsonkeytoken{"node"}@*): "vid1", (*@\jsonkeytoken{"tool"}@*): "generate_video",
          (*@\jsonkeytoken{"first\_frame\_url"}@*): "<workspace>/img_a3f2c1.png" (*@\jsonobjecttoken{\}}@*)
      (*@\jsoninnertoken{\}}@*)
    (*@\jsonobjecttoken{\}}@*)
  (*@\jsonarraytoken{]}@*)
(*@\jsonoutertoken{\}}@*)
\end{jsonlistingbox}
\end{minipage}
\caption{Cross-turn reuse represented as a JSON object.
The second turn regenerates nothing: the referenced asset enters the new graph as a completed node, and only the video node issues a call.}
\label{fig:crossturn}
\end{figure}

Figure~\ref{fig:crossturn} shows the mechanism end to end.
In the first turn a node generates an image and its record enters the registry.
In a later turn, possibly in a different session, the user refers to that image; the host agent locates the record by semantic match over the stored descriptions, and materializes it in the new graph as an asset-source node.
That node is complete on entry, issues no generation call.

\section{Experiment Details}
\label{app:experiment}

\subsection{Construction of UniM-90}
\label{app:experiment:unim90}

UniM-90 is a fixed subset of 90 instances drawn from UniM~\cite{li2026unim}.
We sample one easy, one medium, and one hard case from each of UniM's 30 domains.
If a difficulty level is unavailable, it is replaced by an additional case from another available level within the same domain, yielding 90 cases in total.

\subsection{Configuration}
\label{app:experiment:config}

The \emph{Base Agent} configurations retain only the Skills natively built into Codex and Claude Code, respectively, with no additional Skills installed, whereas the \emph{Agent + Omni-IO Skills} configurations add Omni-IO Skills and its associated MCP tool services while keeping all other settings, including tool configurations, system prompts, and execution budgets, unchanged. The provider bindings in Table~\ref{tab:provider_bindings} are held fixed across all instances and both agents, ensuring that the observed differences are not attributable to different underlying generators. This controlled setup therefore isolates the contribution of Omni-IO Skills.

\end{document}